\documentclass[12pt]{article}
\usepackage{amsmath}
\usepackage{graphicx}
\usepackage{float} 
\usepackage{enumerate}
\usepackage{natbib} 
\usepackage{multirow}

\usepackage{xr}
\makeatletter

\newcommand*{\addFileDependency}[1]{
	\typeout{(#1)}
	\@addtofilelist{#1}
	\IfFileExists{#1}{}{\typeout{No file #1.}}
}\makeatother

\newcommand*{\myexternaldocument}[1]{%
	\externaldocument{#1}%
	\addFileDependency{#1.tex}%
	\addFileDependency{#1.aux}%
}

\myexternaldocument{0-supple}

\newcommand{\blind}{0}

\usepackage{amsmath,amssymb,amsfonts,amsthm,bbm}

\usepackage{mathtools}   
\usepackage{stmaryrd} 

\usepackage{enumitem}   
\usepackage{romannum}

\usepackage[titletoc,title]{appendix}  

\usepackage[table,xcdraw,dvipsnames]{xcolor}   
\usepackage{hyperref}
\newcommand\myshade{85}
\colorlet{mylinkcolor}{YellowOrange}
\colorlet{mycitecolor}{Aquamarine}
\colorlet{myurlcolor}{violet}

\hypersetup{
  linkcolor  = mylinkcolor!\myshade!black,
  citecolor  = mycitecolor!\myshade!black,
  urlcolor   = myurlcolor!\myshade!black,
  colorlinks = true,
}

\usepackage{caption}
\usepackage{subcaption}

\usepackage[normalem]{ulem}
\usepackage{setspace}

\usepackage{graphicx}
\usepackage{comment}
\graphicspath{{figs/}}

\renewcommand{\hat}{\widehat}
\renewcommand{\tilde}{\widetilde}

\newcommand{\bfm}[1]{\ensuremath{\boldsymbol{#1}}} 

\def\bbone{\mathbbm{1}} 

\def\ba{\bfm a}   \def\bA{\bfm A}  
     
\def\bc{\bfm c}     
\def\bd{\bfm d}     
     \def\EE{\mathbb{E}}
    
   \def\bG{\bfm G}

\def\bm{\bfm m}     
     \def\NN{\mathbb{N}}

   \def\bR{\bfm R}  \def\RR{\mathbb{R}}

   \def\bU{\bfm U}  
     
\def\bw{\bfm w}     
\def\bx{\bfm x}     
     
\def\bz{\bfm z}

\def\calB{{\cal  B}} \def\cB{{\cal  B}}
 \def\cC{{\cal  C}}
 \def\cD{{\cal  D}}
 
\def\calF{{\cal  F}} \def\cF{{\cal  F}}
 
 \def\cH{{\cal  H}}
\def\calI{{\cal  I}} \def\cI{{\cal  I}}
 \def\cJ{{\cal  J}}

 \def\cN{{\cal  N}}
\def\calO{{\cal  O}} 
 \def\cP{{\cal  P}}
 
 \def\cR{{\cal  R}}
 \def\cS{{\cal  S}}
\def\calT{{\cal  T}} \def\cT{{\cal  T}}
 \def\cU{{\cal  U}}
 \def\cV{{\cal  V}}
 \def\cW{{\cal  W}}
\def\calX{{\cal  X}} \def\cX{{\cal  X}}

\providecommand{\norm}[1]{\left\lVert#1\right\rVert}

\providecommand{\paren}[1]{\left( #1 \right)}

\providecommand{\defeq}{:=}

\usepackage{mathtools}
\DeclarePairedDelimiterX{\infdivx}[2]{(}{)}{%
  #1 \; \delimsize\| \; #2%
}

\DeclareMathOperator{\rank}{rank}

\DeclareMathOperator{\sgn}{sgn}

\newcommand*\xbar[1]{%
  \hbox{%
    \vbox{%
      \hrule height 0.4pt 
      \kern0.5ex
      \hbox{%
        \kern-0em
        \ensuremath{#1}%
        \kern-0em
      }%
    }%
  }%
} 
\newtheorem{definition}{Definition}

\newtheorem{condition}[definition]{Condition}
\newtheorem{lemma}[definition]{Lemma}
\newtheorem{proposition}[definition]{Proposition}
\newtheorem{theorem}[definition]{Theorem}
\newtheorem{corollary}[definition]{Corollary}

\theoremstyle{definition}
\newtheorem{remark}{Remark}

\definecolor{royalpurple}{rgb}{0.47, 0.32, 0.66}
\definecolor{greenfresh}{HTML}{00897B}
\definecolor{bluefresh}{HTML}{1E88E5}
\definecolor{redfresh}{HTML}{E53935}

\definecolor{royalpurple}{rgb}{0.47, 0.32, 0.66}

\def\beq{\begin{equation}}
\def\eeq{\end{equation}}

\def\bet{\begin{theorem}}
\def\eet{\end{theorem}}

\def\bel{\begin{lemma}}
\def\eel{\end{lemma}}

\def\bcalX{{\boldsymbol{\calX}}}

\usepackage{setspace}
\usepackage{etoolbox}
\BeforeBeginEnvironment{equation*}{\begin{singlespace}\vspace*{-\baselineskip}}
	\AfterEndEnvironment{equation*}{\end{singlespace}\noindent\ignorespaces}
\BeforeBeginEnvironment{equation}{\begin{singlespace}\vspace*{-\baselineskip}}
	\AfterEndEnvironment{equation}{\end{singlespace}\noindent\ignorespaces}
\BeforeBeginEnvironment{align}{\begin{singlespace}\vspace*{-\baselineskip}}
	\AfterEndEnvironment{align}{\end{singlespace}\noindent\ignorespaces}
\BeforeBeginEnvironment{align*}{\begin{singlespace}\vspace*{-\baselineskip}}
	\AfterEndEnvironment{align*}{\end{singlespace}\noindent\ignorespaces}
\BeforeBeginEnvironment{eqnarray}{\begin{singlespace}\vspace*{-\baselineskip}}
	\AfterEndEnvironment{eqnarray}{\end{singlespace}\noindent\ignorespaces}
\BeforeBeginEnvironment{eqnarray*}{\begin{singlespace}\vspace*{-\baselineskip}}
	\AfterEndEnvironment{eqnarray*}{\end{singlespace}\noindent\ignorespaces}

\usepackage[framemethod=TikZ]{mdframed} 

\usepackage{fancybox}

\usepackage[linesnumbered,ruled,vlined]{algorithm2e}

\SetCommentSty{mycommfont}

\SetKwInput{KwInput}{Input}                
\SetKwInput{KwOutput}{Output}              
\SetKwInput{KwInitialize}{Initialize}    

\usepackage{listings}             
\graphicspath{{figs/}}
\begin{document}
\pagenumbering{arabic}

\def\spacingset#1{\renewcommand{\baselinestretch}%
{#1}\small\normalsize} \spacingset{1}

%
%
%

\def\TITLE{Dual-Channel Tensor Neural Networks: Finite-Sample Theory and Conformal Structure Selection}

\if0\blind
{
  \title{\bf \TITLE}
  \author{
    Elynn Chen$^\sharp$\thanks{\scriptsize{Correspondence to E. Chen (E-mail: elynn.chen@stern.nyu.edu) and J. Pei (E-mail: j.pei@duke.edu). }} \hspace{6ex} 
    Jiayu Li$^\diamond$ \hspace{6ex}
    Zheshi Zheng$^\flat$ \hspace{6ex} 
    Jian Pei$^\dag$ \\ 
    \normalsize
    \medskip
    $^{\sharp,\diamond}$New York University \hspace{3ex}
    $^{\flat}$ University of Michigan \hspace{3ex}
    $^\dag$ Duke University
    }
  \maketitle
} \fi

\if1\blind
{
  \bigskip
  \bigskip
  \bigskip
  \begin{center}
    {\LARGE\bf \TITLE}
\end{center}
  \medskip
} \fi


\begin{abstract}
\spacingset{1.08}
\noindent
Tensor-valued data arise naturally in neuroimaging, genomics, climate science, and spatiotemporal networks, where multilinear dependencies across modes carry information that is destroyed under vectorization. Existing approaches either impose a single low-rank structure, which can miss localized signal, or treat the tensor as a long vector, which discards its multiway geometry. We propose a {\em Dual-Channel Tensor Neural Network} (DC-TNN) that decomposes each tensor input into a low-rank core and a sparse refinement, and processes the two components through coupled neural channels. The framework is structure-agnostic and accommodates CP, Tucker, and tensor-train cores within a single architecture. For estimation, we establish non-asymptotic risk bounds for the DC-TNN estimator that decompose into network approximation, core estimation, and refinement-selection terms, and show that the effective dimension is determined jointly by the core rank and refinement sparsity rather than by the ambient tensor size. For inference, we develop a {\em structure-aware conformal ROC} procedure that calibrates within the core-refinement latent space and produces ROC and AUC confidence bands with finite-sample, distribution-free coverage. Building on this, we propose a {\em conformal structure selector} that, to our knowledge, is the {\em first distribution-free procedure} for choosing among candidate tensor decompositions with finite-sample validity. Simulations and an analysis of a protein dataset demonstrate competitive predictive accuracy, reliable uncertainty quantification, and consistent recovery of the tensor structure.
\end{abstract}

\noindent%
\vfill


\newpage
\spacingset{1.9} 

\addtolength{\textheight}{.1in}%

\section{Introduction}  \label{sec:intro}

Modern machine learning applications increasingly involve {\em high-dimensional structured data} that naturally arise as tensors, including multi-modal biomedical measurements, spatiotemporal signals, and graph-structured representations \citep{chen2020modeling,chen2022modeling,wen2024tensor,kong2025teaformers}. Unlike vectorized inputs, tensor data preserve rich {\em multiway dependencies across modes}, which are critical for capturing complex interactions and improving predictive performance \citep{wen2024tensor,kong2025teaformers,wu2025tensor}. A central challenge in learning from such data is that their underlying structure is often {\em heterogeneous}: global patterns governed by low-dimensional latent factors coexist with {\em localized, irregular, or sparse variations} that carry important predictive signals. Developing models that can effectively capture both types of structure remains a fundamental problem in modern data mining and machine learning.

Existing approaches typically address high-dimensional tensor data by imposing {\em either low-rank structure or sparsity}, but each paradigm is fundamentally limited in heterogeneous settings. Low-rank models, including CP, Tucker, and tensor-train representations, are effective at summarizing global multilinear dependencies, yet they often suffer from {\em representation bias} when localized signal components are not aligned with the assumed low-rank subspace. Conversely, sparse or high-dimensional nonparametric models provide flexibility to capture localized effects, but they largely ignore the underlying tensor structure and therefore incur an {\em inflated effective dimension}, leading to reduced statistical and computational efficiency. As a result, neither paradigm alone is sufficient for modeling real-world tensor data that exhibit both global structure and local variability \citep{chen2026factor,chen2025modewise}.

These limitations suggest that the key challenge is not merely the selection of a specific tensor decomposition (e.g., CP versus Tucker), but rather a broader problem of {\em learning under structural uncertainty}. In practice, the true data-generating mechanism may not conform to a single homogeneous structure; instead, it often lies in a richer class that combines {\em low-rank global components with sparse, localized refinements}. This perspective reframes tensor learning as a {\em model class selection problem}, where the goal is to identify and learn an appropriate hybrid structure that balances global regularity and local flexibility. An effective framework should therefore avoid committing to a single structural assumption a priori, and instead enable adaptive integration of multiple structural components within a unified model.
%

To address this challenge, we propose a {\em dual-channel neural architecture} for learning predictive functions over tensor inputs with heterogeneous structure. The key idea is to decompose the input into two complementary components: a {\em low-rank core} capturing dominant multilinear patterns, and a {\em refinement component} encoding localized deviations. These components are processed through parallel network channels and are coupled via {\em cross-channel interactions}, enabling the model to jointly exploit global and local information \citep{wen2024tensor,kong2025teaformers,wen2025bridging}. The resulting framework provides a flexible yet structured inductive bias that generalizes both pure low-rank tensor models and high-dimensional models, allowing the model to adapt to different structural regimes in the data.

Beyond the architectural design, the proposed framework introduces a new {\em structured function class} that bridges low-rank tensor models and high-dimensional sparse models \citep{chen2026factor,chen2024hightensordisc,chen2025high}. By jointly modeling a low-dimensional global component and a sparse refinement component, the framework effectively interpolates between these two regimes, capturing a wider range of data-generating mechanisms. This perspective provides a unified view of tensor modeling and enables a principled analysis of how structural assumptions influence generalization, highlighting the trade-offs between approximation accuracy, structural bias, and model complexity.

In addition to predictive modeling, we incorporate {\em uncertainty quantification} to provide more reliable evaluation and decision-making \citep{wu2024conditionalUQ,wu2025conditional}. Standard conformal prediction methods, when applied to high-dimensional tensor data, typically ignore the underlying structure and treat inputs as unstructured vectors, leading to overly conservative and less informative uncertainty estimates. In contrast, we develop a structure-aware conformal inference procedure that operates on the proposed model, yielding {\em sharper and more informative confidence regions} for performance metrics such as ROC curves and AUC \citep{wu2025conditional}, while maintaining distribution-free validity.

A further innovation of this work is a principled approach to {\em model selection under structural uncertainty}. In practice, choosing among candidate tensor structures (e.g., CP, Tucker) is typically done using heuristic validation procedures that do not account for uncertainty arising from finite samples or model fitting. We address this limitation by developing a statistically grounded selection procedure that compares competing models through their predictive performance while incorporating uncertainty. This allows us to determine whether one structure is significantly better than another, or whether multiple structures are statistically indistinguishable.

The main contributions of this paper are summarized as follows:
\begin{itemize}
\item {\em Structured Modeling Framework.} We propose a dual-channel neural architecture that learns global-local representations for tensor data by jointly modeling low-rank structure and sparse refinements within a unified framework.

\item {\em Unified Modeling Perspective.} We introduce a new function class that bridges low-rank tensor models and high-dimensional sparse models, providing a principled view of representation learning under heterogeneous structural assumptions.

\item {\em Uncertainty Quantification.} We develop a structure-aware conformal inference method that produces sharper and more informative uncertainty estimates for predictive performance.

\item {\em Model Selection under  Structural Uncertainty.} We propose a statistically grounded procedure for selecting among competing tensor structures,  addressing a key gap in existing methods.

\item {\em Theoretical Guarantees.} We establish non-asymptotic generalization bounds and show that the proposed framework achieves optimal rates while explicitly characterizing the trade-offs between approximation accuracy, structural bias, and model complexity in hybrid representation learning.
\end{itemize}

We validate the proposed approach through extensive experiments on synthetic and real-world datasets. The results demonstrate improved predictive performance in heterogeneous settings, effective recovery of underlying structural patterns, and reliable uncertainty quantification and model selection. These findings highlight the practical advantages of combining structured inductive bias with flexible predictive modeling in tensor-based learning problems.

\subsection{Related Work and Our Distinctions}

For clarity we group related work into three threads and state our distinctions explicitly.

\noindent
\textbf{Tensor regression with low-rank, sparse, or hybrid structure.} 
The foundational tensor regression literature establishes CP- and Tucker-based generalized linear models \citep{zhou2013tensor,li2018tucker,chen2024semi,xu2025statistical}, tensor-train regression \citep{si2022efficient}, tensor-response and tensor-on-tensor regression \citep{sun2017provable,li2017parsimonious,lock2018tensor,raskutti2019convex}, Bayesian tensor regression \citep{guhaniyogi2017bayesian}, and provable sparse tensor decomposition \citep{sun2017provable}. 
A related body of work develops matrix- and tensor-variate factor models that exploit multilinear low-rank structure for dimension reduction and prediction \citep{chen2020constrained,chen2023statistical,liu2022identification,liu1904helping,chen2024time}, as well as factor-augmented regression that incorporates latent factors as predictors \citep{chen2026factor}. 
The hybrid linear line, including Tucker-with-sparsity \citep{ahmed2020tensor}, boosted sparse CP \citep{he2018boosted}, cubic sketching for sparse-and-low-rank tensors \citep{hao2020sparse}, and the generalized low-rank-plus-sparse estimator of \citep{cai2023generalized}, considers both low-rank and sparse components. 
Tensor classification and discriminant analysis under CP and Tucker low-rank structures \citep{chen2024hightensorclass,chen2024hightensordisc,chen2025high,liu2025tensor} further illustrate the value of exploiting tensor structure in supervised learning. 
The non-asymptotic theory of generalized tensor estimation is developed in \citep{han2022optimal}, with recent inference advances for low-rank tensor models in \citep{xu2025statistical}.

{\em We depart from this body of work in two respects.} First, where these methods impose low-rank-plus-sparse structure on a {\em linear} coefficient tensor, we model the regression function as a {\em nonlinear} map on the latent representation $(\mathcal{C}, \mathcal{V}_{\mathcal{J}}) $, producing a strictly richer function class. Second, we provide finite-sample distribution-free inference and a structure-selection procedure, neither of which exists for the linear estimators.

\smallskip\noindent
\textbf{Neural networks for tensor inputs and for nonparametric regression.} 
On the architectural side, weight-tensorization methods \citep{novikov2015tensorizing,kossaifi2020tensor} reduce parameter counts but do not model input structure, do not separate global from local components, and do not provide non-asymptotic statistical guarantees. 
Tensor-contraction and tensor-augmented architectures \citep{wen2024tensor,wu2024conditionalUQ,wu2025tensor,wu2025conditional,kong2025teaformers,wen2025bridging} preserve multilinear structure in activations but operate with a single, fixed structural assumption. 
On the theoretical side, our analysis builds on the analysis for deep ReLU networks under Hölder smoothness \citep{schmidthieber2020nonparametric,bauer2019deep, kohler2021rate}, revised here in the tensor-input setting and combined with tensor-factor-model error propagation \citep{zhang2018tensor,chen2024hightensorclass}. 

The closest methodological predecessor is FAST-NN \citep{fan2023factor}, which establishes the latent-factor-plus-sparse-idiosyncratic paradigm for deep ReLU networks with vector covariates; our framework lifts that paradigm to tensor covariates and substantially extends it through structure-aware inference and tensor structure selection.

\smallskip\noindent
\textbf{Conformal inference for classification and ROC curves.}
Split conformal prediction supplies distribution-free intervals and prediction sets \citep{vovk2005algorithmic,shafer2008tutorial,lei2018distribution}, with extensions to heteroscedastic regression \citep{romano2019conformalized} and localized calibration \citep{guan2023localized}. 
For classification, recent work develops conformal ROC and AUC inference \citep{zheng2024quantifying,zheng2025classification} and adapts conformal ideas to tensorized graph models \citep{wu2024conditionalUQ,wu2025conditional}. Applied to tensor inputs, these methods calibrate in the ambient space and yield conservative bands. Our procedure calibrates in the DC-TNN's latent core-refinement representation, producing sharper bands while retaining distribution-free coverage, and extends to the difference-ROC setting that underlies our structure selector. To our best knowledge, this is the first prediction-based test for low-rank tensor structure selection with finite-sample validity.

\smallskip\noindent
\textbf{Organization.} 
Section 2 formalizes the core–refinement model and the DC-TNN architecture. Section 3 details the two-stage training procedure. Section 4 develops the finite-sample theory. Section 5 develops the structure-aware conformal inference procedure. Section 6 introduces the conformal structure selector. Sections 7 and 8 report simulation and real-data results on the DD protein benchmark. Section 9 concludes. Proofs, the end-to-end training procedure, the tensor-train variant, and additional experiments are deferred to the supplement.

\section{Structured Modeling of Heterogeneous Tensor Data} \label{sec:structured_modeling}

We consider learning predictive models from {\em high-dimensional tensor inputs}, where each observation is a multiway array $\cX \in \RR^{D_1 \times \cdots \times D_M}$. A key challenge is that real-world tensors often exhibit {\em heterogeneous structure}, where global low-rank patterns coexist with {\em localized and irregular variations} that are critical for prediction.

\subsection{Structured Feature Transformation}
To address this challenge, we consider a decomposition of the form
\begin{equation}\label{eq:core_refine}
\cX = \cS(\cC) + \cV,
\end{equation}
where $\cC \in \RR^{R_1 \times \cdots \times R_M}$ is a latent core tensor with $R_m \ll D_m$, and $\cS: \RR^{R_1 \times \cdots \times R_M} \rightarrow \RR^{D_1 \times \cdots \times D_M}$ is a multilinear map that expands $\cC$ into the ambient tensor space. The term $\cS(\cC)$ represents a {\em global low-rank component}, encoding dominant multilinear interactions, while $\cV$ captures {\em localized or idiosyncratic variations} not explained by the global structure.

Importantly, this decomposition is not imposed as a generative model, but defines a {\em structured feature transformation}: 
$\cX \mapsto (\cC, \cV)$. Since $\cV$ captures localized and irregular deviations from the global structure, it is expected to be sparse, only a small subset of its entries carry predictive information. This motivates restricting to local features $\cV_{\cJ}$, the entries of $\cV$ indexed by $\cJ$, reducing the transformation to $\cX \mapsto (\cC, \cV_{\cJ})$. The index set $\cJ$ is determined by a sparsity-constrained selection procedure detailed in Section \ref{subsec:training}. This transformation serves as a form of {\em structured inductive bias}, rather than a constraint on the predictive model itself. The predictive function is then modeled as a {\em nonlinear mapping over $(\cC, \cV_{\cJ})$}, integrating {\em low-rank global structure with sparse local flexibility}, and effectively reducing complexity compared to operating directly in the ambient tensor space.

\subsection{Dual-Channel Neural Architecture} \label{sec:architecture}

Building on the structured feature transformation, we propose a {\em dual-channel neural architecture} that jointly models global and local components by processing $\cC$ and $\cV_J$ through {\em separate but interacting channels}. The {\em core channel} $\cH_c$ operates on the low-dimensional core tensor $\cC$, learning representations that capture global multilinear dependencies. In parallel, the {\em refinement channel} $\cH_u$ processes $\cU \in \RR^{K_1\times\cdots\times K_M}$, a dense tensor induced by $\cV$, storing the selected entries of the local component $\cV$ indexed by $\cJ$, with dimensions $K_1,\ldots,K_M$ chosen such that $\prod_{m=1}^M K_m \geq |\cJ|$ and approximately balanced ($K_1 \asymp \cdots\asymp K_M$), ensuring sufficient capacity to represent all active refinement locations. The two channels are coupled through {\em cross-channel interactions}, allowing information to flow between global and local representations at each layer, enabling the model to adaptively combine structured and flexible components.

We initialize $\cH_{c}^{(0)} = \cC$ and $\cH_u^{(0)} = \cU$. At each layer $\ell$, let $\cH_c^{(\ell)} \in \RR^{R_1^{(\ell)} \times \cdots \times R_M^{(\ell)}}$ and $\cH_u^{(\ell)} \in \RR^{K_1^{(\ell)} \times \cdots \times K_M^{(\ell)}}$ denote the hidden states of the core and refinement channels. The layerwise updates are given by
\begin{equation} \label{eqn:TTL-FA}
	\begin{aligned}
		\text{Core ($\cC$) channel:}\quad \cH_{c}^{(\ell+1)}& \defeq \alpha_c\paren{ \cW^{(\ell)}_{cc}\bullet\cH_{c}^{(\ell)} + \cW^{(\ell)}_{cu}\bullet\cH_{u}^{(\ell)} + \cB_{c}^{(\ell)} }, \\
		\text{Refinement ($\cU$) channel:}\quad \cH_u^{(\ell+1)}& \defeq \alpha_u\paren{ \cW^{(\ell)}_{uc}\bullet\cH_{c}^{(\ell)} + \cW^{(\ell)}_{uu}\bullet\cH_{u}^{(\ell)} + \cB_u^{(\ell)} },
	\end{aligned}
\end{equation}
where $\alpha_c,\alpha_u$ are nonlinear activation functions applied elementwise, and $\bullet$ denotes tensor contraction along matching modes. The weight tensors $(\cW^{(\ell)}_{cc},\cW^{(\ell)}_{uu})$ learn structure within each channel, while $(\cW^{(\ell)}_{cu},\cW^{(\ell)}_{uc})$ model the interactions between the core and refinement representations. The bias tensors $\cB_c^{(\ell)} \in \RR^{\times_{m=1}^M R_m^{(\ell+1)}}$ and $\cB_u^{(\ell)} \in \RR^{\times_{m=1}^M K_m^{(\ell+1)}}$ account for affine shifts.

Denote by $\calT^{(\ell+1)}(\cdot,\cdot)$ the parallel update operator in \eqref{eqn:TTL-FA}, which maps $\paren{\cH_c^{(\ell)}, \cH_u^{(\ell)}} \rightarrow \paren{\cH_c^{(\ell+1)}, \cH_u^{(\ell+1)}},\; \ell=0,\ldots,L-1$. After $L$ hidden layers, we apply a final linear operator 
\[
\cT^{(L+1)}(\cH_c^{(L)}, \cH_u^{(L)}) = \cW_c^{(L)} \bullet \cH_c^{(L)} + \cW_u^{(L)} \bullet \cH_u^{(L)} + \cB^{(L)},
\]
with $\cW_{c}^{(L)} \in \RR^{d^{(L+1)}\times_{m=1}^M R_m^{(L)}}$, $\cW_{u}^{(L)} \in \RR^{d^{(L+1)}\times_{m=1}^M K_m^{(L)}}$,  
and the bias $\cB^{(L)} \in \RR^{d^{(L+1)}}$.
The resulting output is a $d^{(L+1)}$-dimensional vector (with $d^{(L+1)}=1$ for scalar regression, and $d^{(L+1)}=K-1$ for $K$-category classification). Thus, the dual-channel tensor neural network defines the mapping
\begin{equation}
\label{eqn: dual-channel TNN}
f(\cC,\cU) = \alpha\!\left(\calT^{(L+1)} \circ\calT^{(L)}\circ \cdots\circ\calT^{(1)}(\cC,\cU)\right),
\end{equation}
where $\alpha$ denotes the link function (identity for regression, logistic or softmax for classification). The operator in \eqref{eqn: dual-channel TNN} defines a class of dual-channel tensor neural networks, formalized in Definition \ref{def: dual deep relu} and illustrated in Figure 1.
\begin{definition}[Dual-Channel Deep ReLU Tensor Network]
	\label{def: dual deep relu}
	~\\
	Let $\bd^{(\ell)}=(R_1^{(\ell)}, \dots, R_M^{(\ell)}, K_1^{(\ell)}, \dots, K_M^{(\ell)})$ denote a $2M$-dimensional width vector at layer $\ell$, where the first $M$ entries specify the widths of the core channel and the remaining $M$ entries specify those of the refinement channel.  
	Define the width tuple $\bd=(\bd^{(1)}, \dots, \bd^{(L)}, d^{(L+1)})$, where $d^{(L+1)}\in\NN$ is the output width of the final layer.
	
	\noindent For any depth $L\in\NN$, width tuple $\bd$, truncation level $V \in \RR^+ \cup \{\infty\}$, and weight bound $B \in \RR^+$, the class of truncated dual-channel deep ReLU tensor networks is defined as
	\begin{equation*}
		\calF(L, \bd, V, B) = \left\{\bar{f}(\cC, \cU) = \calT_{V}(f(\cC, \cU))\} \right.,  
	\end{equation*}
	where $f(\cC, \cU)$ is the DC-TNN mapping defined in \eqref{eqn: dual-channel TNN} with parameters satisfying
	\begin{equation*}
		\max_{1\le \ell \le L} \big\{\norm{\cW_{cc}^{(l)}}_{\max}, \norm{\cW_{cu}^{(l)}}_{\max}, \norm{\cW_{uu}^{(l)}}_{\max}, \norm{\cW_{uc}^{(l)}}_{\max},\; \norm{\calB_u^{(\ell)}}_{\max}, \norm{\calB_c^{(\ell)}}_{\max} \big\} \le B,
	\end{equation*}
    and the output-layer parameters satisfy $\|\cW_c^{(L)}\|_{\max}, \|\cW_u^{(L)}\|_{\max}, \|\cB^{(L)}\|_{\max} \leq B$.
Here, $\calT_{V}(\cdot)$ denotes elementwise truncation at level $V$: $[\calT_{V}(z)] = \sgn(z)( |z| \wedge V)$ applied to each coordinate of the output. For brevity, when the width tuple is of the form $\bd = (\bd_{in}, \bw, \cdots, \bw, d_{out})$, we write $\calF(L, \bd_{in}, d_{out}, \bw, V, B)$.
\end{definition}

\begin{remark}
	The architecture in Definition~\ref{def: dual deep relu} encompasses several important special cases in tensor learning. When all cross- and refinement-channel weights vanish, i.e.,  
	$\cW_{uc}^{(\ell)}=\cW_{cu}^{(\ell)}=\cW_{uu}^{(\ell)}=0$ for all $\ell$, the network reduces to a tensor neural network operating solely on $\cC$, corresponding to a purely low-rank model. Conversely, when all core-related weights vanish, i.e., $\cW_{uc}^{(\ell)} =\cW_{cu}^{(\ell)} =\cW_{cc}^{(\ell)}=0$, the network operates solely on $\cU$, which can be viewed as a sparsity-constrained model.
	More generally, the dual-channel formulation unifies these two complementary mechanisms: the core channel captures structured multilinear interactions through a low-dimensional latent tensor, while the refinement channel provides localized, sparsity-constrained flexibility. Together, they enable richer representations than either mechanism alone.
\end{remark}

\subsection{Model Instantiations and Structural Variants} 

The framework in \eqref{eq:core_refine} can be instantiated using different choices of the signal map $\cS(\cdot)$, including Tucker, and CP decompositions, each inducing a distinct structural bias on the global component $\cC$. Since in practice, the appropriate structure is often unknown and depends on the underlying data, we address {\em data-driven selection among candidate structures}through a principled model selection procedure developed in Section~\ref{sec:selector}.

\medskip\noindent
\textbf{Tucker-Core Adaptation.}  
Each tensor covariate $\cX_i \in \RR^{\times_{m=1}^M D_m}$ is decomposed as
\begin{equation}
	\label{eqn:tnn-tucker}
	\cX_i = S(\cC_i) + \cV_i = \cC_i \times_{m=1}^M \bA_m + \cV_i, \quad i \in [n],
\end{equation}
where $\cC_i \in \RR^{\times_{m=1}^M R_m}$ is the latent Tucker core, $\bA_m \in \RR^{D_m \times R_m}$ are the mode-$m$ loading matrices with $R_m \ll D_m$, and $\cV_i$ captures variation not explained by the low-rank component. This representation preserves the multiway structure of the original covariate, enabling the core channel to learn global multilinear dependencies directly from a low-dimensional latent tensor, and is well-suited when the tensor modes exhibit rich, simultaneous interactions.


\medskip\noindent
\textbf{CP-Core Adaptation.}  
The CP variant decomposes $\cX_i$ as
\begin{equation}
	\label{eqn:cp decomposition}
	\cX_i = S(\cC_i) + \cV_i = \sum_{r=1}^R c_{i,r}\, \ba_{1r} \circ \ba_{2r} \circ \cdots \circ \ba_{Mr} + \cV_i,
\end{equation}
where $\circ$ denotes the outer product, $\{\ba_{mr} \in \RR^{D_m}\}_{r=1}^R$ are unit-norm mode-$m$ factor vectors, and the latent core $\cC_i$ is represented by the coefficients $\{c_{i,r}\}_{r=1}^R$ via a super-diagonal tensor core whose nonzero entries correspond to these coefficients. The CP core is more parsimonious than Tucker's core. Unlike Tucker, CP does not require orthogonal loading matrices, allowing the factor vectors to be collinear, which provides greater flexibility and interpretability as each component corresponds to a rank-1 outer product contribution.


\medskip\noindent
\textbf{Tensor-Train Core Adaptation.}  
The Tensor-Train (TT) variant \citep{TensorTrain} offers an alternative instantiation where $\cX_i$ is expressed as a sequence of third-order TT cores:
\begin{equation}
	\label{eqn:tt decomposition}
	\cX_i(i_1,\dots,i_M) = \cC_i^{(1)}(1,i_1,:) \; \cC_i^{(2)}(:,i_2,:) \cdots \cC_i^{(M-1)}(:,i_{M-1},:) \; \cC_i^{(M)}(:,i_M,1) + \cV_i,
\end{equation}
where $\cC_i^{(m)} \in \RR^{R_{m-1} \times D_m \times R_m}$ denotes the $m$-th TT core with boundary conditions $R_0 = R_M = 1$. The collection of TT cores $\cC_i := \{ \cC_i^{(1)},\ldots,\cC_i^{(M)}\}$ is processed through \emph{embedded tensor-train layers (E-TTL)}, a multi-channel extension of the DC-TNN framework, where the refinement representation $\cU_i$ induced by $\cV_i$ is incorporated as an additional TT core. This augmented TT chain enables interactions both among TT cores and between the core and refinement component, while preserving the TT-rank structure. The sequential chain structure naturally captures hierarchical dependencies between modes. The detailed E-TTL architecture is presented in Appendix~\ref{sec:ETTL}.

\medskip

Together, the Tucker, CP, and TT adaptations cover a broad spectrum of low-rank tensor structures, from multilinear factorizations to sequential tensor-train representations, providing a unified framework where the core channel captures structured global interactions and the refinement channel provides complementary sparse flexibility. The appropriate adaptation is selected via the data-driven procedure developed in Section~\ref{sec:selector}.

\section{Training of Dual-Channel Tensor Neural Networks}
\label{subsec:training}

Two training procedures are developed for the DC-TNN framework, with the two-stage procedure presented in this section and an alternative end-to-end procedure in Appendix \ref{append:end-to-end} (Tucker and CP) and Appendix \ref{sec:ETTL} (TT). In the two-stage procedure, the core tensor $\widetilde{\cC}_i$ is first estimated outside the network by projecting $\cX_i$ onto estimated loading matrices via the inverse of $\cS(\cdot)$, and then fixed during network training. This separation is adopted for theoretical convenience: under mild regularity conditions, standard estimators such as HOSVD \citep{zhang2018tensor} for Tucker and ALS \citep{anandkumar2015} for CP, achieve consistent core estimation, making it analytically tractable to characterize how estimation error propagates through the dual-channel network, as formalized in Theorem~\ref{thm:optimal_inequality}. In the end-to-end procedure, the decomposition parameters defining $\cS(\cdot)$, is jointly optimized with the network weights and sparse refinement selector in a single training pipeline. Proposition \ref{pro:equivalence under the linear truth} in Appendix \ref{append:end-to-end} further establishes that with warm start and a local stability condition, the two-stage and end-to-end procedures are locally asymptotically equivalent.

In the two-stage procedure, the core tensor $\widetilde{\cC}_i$ is estimated from $\cX_i$ via the estimated inverse of the signal map $\hat\cS^{-1}(\cdot)$, which admits a closed-form expression involving learned projection operators for Tucker and CP decompositions. With $\widetilde{\cC}_i$ fixed, the network parameters and refinement selector $\cW_u$ are jointly optimized. Rather than estimating the refinement component as the residual $\cX_i-\hat\cS(\widetilde{\cC}_i)$, which is high-dimensional and sensitive to core estimation error, making error propagation through the network analytically challenging, we instead obtain the refinement representation directly via $\cU_i=\cW_u \bullet \cX_i$. Here $\cW_u$ is a learned selection tensor with sparsity constraints that extracts a sparse and low-dimensional refinement input directly from $\cX_i$. This design also makes the interaction between the core and refinement components explicit: while $\widetilde{C}_i$ captures the global low-rank structure, $\cW_u$ implicitly models the complementary local structure by learning which coordinates of $\cX_i$ carry predictive information beyond what the core explains. We define the approximate effective dimension as $\bd_{\mathrm{in}} = (\bar{R}_1,\ldots,\bar{R}_M,\bar{K}_1,\ldots,\bar{K}_M)$, where $\bar{R}_m \ge R_m$ are possibly over-specified ranks and $\bar{K}_m$ control the dimensionality of the refinement component. The joint optimization solves
\begin{equation} 
	\label{eqn:TTN-penalty}
	\begin{split}
		\hat{f}_n,\; \hat{\cW}_u &\in \underset{\substack{f \in \calF(L, \bd_{\text{in}}, 1, \bd_{\text{app}}, V, B) \\ \cW_u \in \RR^{\times_{m=1}^M \bar K_m \times_{m=1}^M D_m} }}{\arg\min} \left(\cR_n(f, \cW_u) + \cP_n(\cW_u;\rho_{\lambda})\right), \\
	\cR_n(f,\cW_u) = \frac{1}{n} \sum_{i=1}^n & \left(y_i - f(\tilde \cC_i, \cW_u \bullet \cX_i \right)^2, \; \cP_n(\cW_u;\rho_{\lambda}) = \sum_{\calI=(i_1,\dots, i_{2M})} \rho_{\lambda}\paren{(\cW_{u})_{\calI}},
\end{split}
\end{equation}
where the hidden layer widths are all the same with $\bd_{\mathrm{app}} = (\tilde{R}_1,\ldots,\tilde{R}_M,\tilde{K}_1,\ldots,\tilde{K}_M)$, and $\rho_{\lambda}:\RR \rightarrow [0, \infty)$ is a coordinate-wise penalty (e.g., $\ell_1$, clipped-$L_1$, lasso, SCAD). Throughout the paper, the default choice is clipped-$L_1$ with the clipping threshold $\tau>0$:
\[\rho_{\lambda}(\cX)=\lambda \psi_{\tau}(\cX), \;\; \psi_{\tau}(\cX) = \frac{|x|}{\tau} \wedge 1.
\]
\textbf{Core estimation error propagation.}
Since $\widetilde{\cC}_i$ is a plug-in estimate of $\cC_i$, the core estimation error propagates into the dual-channel learning procedure. We focus on the Tucker decomposition; the CP case follows analogously. For Tucker, the loading matrices $\{\bA_m\}_{m=1}^M$ are first estimated as $\{\widehat{\bA}_m \in \RR^{\bar R_m \times D_m}\}_{m=1}^M$ via HOSVD, and the core is then estimated by projecting $\cX_i$ onto these estimated loadings: $\widetilde{\cC}_i=\cX_i\times_{m=1}^M \widehat{\bA}_m^\top$. Substituting $\cX_i=\cC_i\times_{m=1}^M \bA_m+\cV_i$ gives
\[
\widetilde{\cC}_i
= \bigl(\cC_i\times_{m=1}^M \bA_m+\cV_i\bigr)\times_{m=1}^M \widehat{\bA}_m^\top
= \cC_i\times_{m=1}^M \bigl(\widehat{\bA}_m^\top \bA_m\bigr)
+ \cV_i\times_{m=1}^M \widehat{\bA}_m^\top.
\]
Setting $\bG_m=\widehat{\bA}_m^\top \bA_m\in\mathbb{R}^{\bar{R}_m\times R_m}$ and $\Xi=\cV_i\times_{m=1}^M \widehat{\bA}_m^\top$, we obtain $\widetilde{\cC}=\cC\times_1^M \bG_m+\Xi$, where $\Xi$ captures the projection of the local residual $\cV_i$ onto the estimated loading matrices. By the Lipschitzness of $f$ and $\widetilde{\cC}\times_{m=1}^M \bG_m^\dagger - \cC = \Xi\times_{m=1}^M \bG_m^\dagger$, where $\bG_m^\dagger$ is left inverse of $\bG_m$,
\[
\bigl|f(\widetilde{\cC},\cU)-f(\cC,\cU)\bigr|
\lesssim
\left\|\widetilde{\cC}\times_{m=1}^M \bG_m^\dagger - \cC\right\|_F
=
\left\|\Xi\times_{m=1}^M \bG_m^\dagger\right\|_F
\le
\prod_{m=1}^M \sigma_{\min}(\bG_m)^{-1}\cdot \|\Xi\|_F.
\]
A formal bound on $\mathbb{E}[\|\Xi\|_F^2]$ is established in Theorem~\ref{thm:optimal_inequality}.

\begin{algorithm}[htbp!]
\caption{Two-Stage Training of Dual-Channel Tensor Neural Networks}\label{algo:FATNN}
\DontPrintSemicolon
\KwInput{Training set $\{(\cX_i,y_i)\}_{i=1}^n$; decomposition type \texttt{type}$\in\{\mathrm{Tucker},\mathrm{CP}\}$; network depth $L$; layer widths $\{\bd^{(\ell)}\}_{\ell=1}^{L}\cup\{d^{(L+1)}\}$; truncation level $V$; Tucker/CP ranks $\{\bar R_m\}_{m=1}^M/ \bar R$; refinement dimensions $\{\bar K_m\}_{m=1}^M$; penalty $\rho_{\lambda}$; activation function $\alpha$}
\KwOutput{Learned predictor $\widehat f_n$, selector $\widehat{\cW}_u$, and residuals $\{\widehat r_i\}_{i=1}^n$}

Compute core representations $\widetilde{\cC}_i=\hat \cS^{-1}(\cX_i)$ for all $i\in[n]$ as follows\;

\Indp
\If{\texttt{type} = Tucker}{
    Compute orthonormal projection matrices $\{\widehat{\bA}_m\}_{m=1}^M$ via HOSVD\;
    Compute cores $\widetilde{\cC}_i=\cX_i\times_{m=1}^M \widehat{\bA}_m^\top$ for all $i\in[n]$\;
}
\ElseIf{\texttt{type} = CP}{
    Run ALS to obtain normalized factor matrices $\{\widehat{\bA}_m\in\mathbb{R}^{d_m\times \bar R}\}_{m=1}^M$\;
    Compute the coefficient vector $\widehat{\bc}_i\in\mathbb{R}^{\bar R}$ by $\widehat{\bc}_i=\widehat{\bG}^{-1}\widehat{\bm}_i$, where $\widehat{\bm}_i(r)=\left\langle \cX_i,\widehat{\ba}_{1r}\otimes\cdots\otimes\widehat{\ba}_{Mr}\right\rangle,
    \; \widehat{\bG}=*_{m=1}^M\bigl(\widehat{\bA}_m^\top\widehat{\bA}_m\bigr)$,
    with $*$ denoting the Hadamard product\;
    Set $\widetilde{\cC}_i\leftarrow \widehat{\bc}_i$ for all $i\in[n]$\;
}
\Indm

Obtain $(\widehat f_n,\widehat{\cW}_u)$ by solving
\[
(\widehat f_n,\widehat{\cW}_u)
\in
\arg\min_{f\in\cF(L, \{\bd^{(\ell)}\}_{\ell=1}^{L}\cup\{d^{(L+1)}\},V,B),\ \cW_u\in\mathbb{R}^{\bar K_1\times\cdots\times \bar K_M\times \prod_{m=1}^M D_m}}
\left\{
\cR_n(f,\cW_u)+\cP_n(\cW_u;\rho_\lambda)
\right\},
\]
where $\cR_n(f,\cW_u)$ and $\cP_n(\cW_u;\rho_\lambda)$ are defined in \eqref{eqn:TTN-penalty}\;

Record residuals $\widehat r_i\leftarrow \widehat y_i-y_i$ for each $i\in[n]$\;

\Return{$\widehat f_n$, $\widehat{\cW}_u$, and $\{\widehat r_i\}_{i=1}^n$}\;
\end{algorithm}

\section{Statistical Guarantees for Prediction}
\label{sec:convergence}


This section establishes finite-sample risk bounds for the DC-TNN estimator. We show that under the core-refinement formulation, the estimator achieves optimal nonparametric rates, where the effective dimension depends on both the low-rank core and sparse refinement components. The resulting error decomposes into three interpretable terms: (i) approximation error of the neural network class, (ii) error propagated from estimating the latent low-rank structure, and (iii) complexity from sparse refinement selection. Together, these terms highlight how the dual-channel design reduces the intrinsic dimension of the problem compared to models relying on a single structural assumption.

\subsection{Model Formulation and Assumptions}
\label{sec:tensor-augmented regression}



Let $\cX \in \RR^{\times_{m=1}^M D_m}$ denote a tensor covariate and $y\in \mathbb{R}$ (or $\{0,1\}$ for binary classification) the response. The general regression function $f^*(\cX)=\mathbb{E}[y\mid \cX]$ minimizes the population $L^2$ risk $R(f)=\mathbb{E}\bigl[(y-f(\cX))^2\bigr]=\int |y-f(\cX)|^2 \,\mu(d\cX,dy)$, but directly learning $f^*$ from $\cX$ is challenging due to the high ambient dimension $D_{\mathrm{tot}}=\prod_{m=1}^M D_m$.
We instead rewrite the model in the \emph{core-refinement form}
\[
y=f^*(\cC,\cV_J)+\varepsilon,
\qquad
\mathbb{E}[\varepsilon\mid \cC,\cV_{\cJ}]=0,
\]
where $\cC$ captures global low-rank structure and the response depends on $\cV$ only through the sparse index set $\cJ\subset [D_1]\times \cdots \times [D_M]$, so that $\cV_{\cJ}$ encodes the localized variations relevant for prediction.

We impose the following assumptions, which are standard in tensor analysis and nonparametric regression, and ensure regularity of the learning problem. We focus on the Tucker instantiation; the analogous conditions for CP are provided in Appendix~\ref{append: assumption}.

\noindent \textbf{Structural Assumptions (Tensor Decomposition).}
We assume the signal map $\cS(\cdot)$ corresponds to a standard low-rank tensor factorization (Tucker or CP), with well-conditioned loading matrices and bounded latent core, ensuring the multilinear mapping is stable and invertible up to affine transformations. When the true ranks are known, the latent core can be \emph{consistently recovered} (up to affine transformation) via HOSVD or HOOI. In practice, the true ranks are often unknown; Condition \ref{cond: sig strength}(iv) allows mild over-parameterization by working with $\bar R_m \geq R_m$, at the cost of an additional residual $\Xi$ that vanishes as the ambient dimensions grow, as established in Theorem \ref{thm:optimal_inequality}. The formal conditions are stated below.



\begin{condition}[Tensor Factor Model] \label{cond: sig strength} For the Tucker factor model \eqref{eqn:tnn-tucker}, we assume:

\noindent
(i)[\textbf{Well-conditioned loadings.}] There exist constants $c_{+} \ge c_{-}>0$ such that the largest and the smallest singular values of each loading matrix $\bA_m$ satisfy $c_{-}D_m^{1/2} \le \sigma_{min}(\bA_m) \le \sigma_{1}(\bA_m) \le c_{+}D_m^{1/2}$.

\noindent
(ii)[\textbf{Core recovery.}] If the true ranks $\{R_m\}_{m=1}^M$ are used, the estimated orthonormal loading matrices $\widetilde{\bA}_m \in \mathbb{R}^{D_m \times R_m}$ obtained via HOSVD span the true loading subspaces with high probability, and the latent core $\cC$ can be consistently recovered up to affine transformation via $\widetilde{\cC} = \prod_{m=1}^M \cX \times_{m=1}^M \widetilde{\bA}_m^\top$.

\noindent
(iii)[\textbf{Bounded core.}] The latent core tensor satisfies $\norm{\cC}_F = \calO(1)$.

\noindent
(iv)[\textbf{Over-parameterization.}] For each mode $m$, choose $\bar R_m \geq R_m$ and let $\hat \bA_m \in \mathbb{R}^{D_m \times \bar R_m}$ be a semi-orthogonal matrix obtained via HOSVD. Define $\bG_m = \hat\bA_m^\top \bA_m \in \mathbb{R}^{\bar R_m \times R_m}$, then $\rank(\bG_m)=R_m$ and $c_- \leq \sigma_{\min}(\bG_m) \leq \sigma_1(\bG_m) \leq c_+$.
\end{condition}

\begin{remark}
Conditions \ref{cond: sig strength} (i) and \ref{cond: sig strength} (iii) are standard in tensor decomposition literature \citep{zhang2018tensor,chen2024highdimensional}. We directly impose Condition \ref{cond: sig strength} (ii) which ensures the signal term $\prod_{m=1}^M \cC\times_{m=1}^M (\bA_m\tilde \bA_m^\top)$ dominates the noise term $\prod_{m=1}^M \cV\times_{m=1}^M \tilde \bA_m^\top$ so the latent core can be reliably recovered. Equivalently, one could assume a signal-to-noise ratio and apply the Davis-Kahan theorem, but we omit this for simplicity. 
\end{remark}

\smallskip
\noindent
\textbf{Smoothness of the Regression Function.} We assume the true regression function $f^*$ is bounded and smooth over the space $(\cC, \cV_{\cJ})$, where the {\em effective dimension} is
$d_{\text{eff}} = \prod_{m=1}^M R_m + |\cJ|$. Condition \ref{cond:reg-func} specifies boundedness and Lipschitz continuity of $f^*$, which controls its global behavior; Condition \ref{cond:Holder} specifies H\"{o}lder smoothness, which controls the local regularity of its higher-order derivatives and enables approximation by neural networks.
\begin{condition}[Regression function] 
	\label{cond:reg-func}
	The true regression function $f^*$ satisfies $\norm{f^*}_\infty\le V^*$ and $m^*$ is $c$-Lipschitz for some universal constants $V^*$ and $c$. We further assume that $1\le V^*\le V \le c'V^*$
	for some universal constant $c'>1$.
\end{condition}

\begin{condition}[H\"{o}lder smoothness of $f^*$]
\label{cond:Holder}
The true regression function $f^*$ belongs to a $(\beta, C)$-H\"{o}lder class with effective dimension $d_{\text{eff}}$. Specifically, let $\beta = a + b$ for some nonnegative integer $a$ and $0 < b \leq 1$, and $C > 0$. For every multi–index $\boldsymbol{\alpha} \in \mathbb{N}^d$ such that $\sum_{j=1}^{d_{\text{eff}}} \alpha_j = r$, the partial derivative $(\partial f^*)/(\partial \bx_1^{\alpha_1} \cdots \partial \bx_{d_{\text{eff}}}^{\alpha_{d_{\text{eff}}}})$ exists and satisfies
\begin{equation*}
\left| \frac{\partial^r f^*}{\partial \bx_1^{\alpha_1} \cdots \partial \bx_{d_{\text{eff}}}^{\alpha_{d_{\text{eff}}}}}(\bz) - \frac{\partial^r f^*}{\partial \bx_1^{\alpha_1} \cdots \partial \bx_{d_{\text{eff}}}^{\alpha_{d_{\text{eff}}}}}(\bz') \right| \leq C \|\bz - \bz'\|_2^s,
\end{equation*}
where $\bx=(\text{vec}(\cC)^\top, \text{vec}(\cV_{\cJ})^\top)^\top$.
\end{condition}

\smallskip
\noindent
\textbf{Regularity Conditions (Boundedness, Dependence, and Noise).} 
We assume the entries of both the core tensor $\cC$ and refinement component $\cV_J$ are uniformly bounded, the refinement component satisfies a weak dependence condition across tensor entries, the noise $\varepsilon$ is sub-Gaussian conditional on $(\cC,\cV_J)$, and $\cC$ and $\cV_J$ satisfy a weak dependence condition between each other. These assumptions are standard in high-dimensional nonparametric regression and ensure concentration of empirical processes and stability of estimation. The detailed conditions are provided in Appendix \ref{append: assumption}.

Under these assumptions, we proceed to establish finite-sample guarantees for the DC-TNN estimator.

\subsection{Finite-Sample Risk Bounds for DC-TNN}

We now establish finite-sample guarantees for the DC-TNN estimator. Let $\hat{f}(\cX)$ denote the DC-TNN estimator obtained from the two-stage training procedure in Section \ref{subsec:training}. Under the assumptions in Section \ref{sec:tensor-augmented regression}, the following finite-sample risk bound holds.

\begin{theorem}[Optimal rate for the DC-TNN estimator]
\label{thm:optimal_inequality}
Work in the Tucker DC-TNN model under Conditions \ref{cond: sig strength}-\ref{cond:Holder} and \ref{cond: boundedness}-\ref{cond:sub-gaussian}. Let $\calF(L, \bd_{\text{in}}, 1, \bd_{\text{app}}, V, B)$ be the DC-TNN class with $\bd_{\text{app}} = (\tilde R_1, \ldots, \tilde R_M, \tilde K_1, \ldots, \tilde K_M)$. We further denote $W=\prod_{m=1}^M \tilde R_m+ \prod_{m=1}^M \tilde K_m$ and $Q=(\prod_m \bar K_m)(\prod_m D_m)$. Let $\hat{f}$ and $\hat \cW_u$ be any $\delta_{\text{opt}}$-approximate empirical risk minimizers
\begin{equation*}
\hat{f}\paren{\cX;\; [\tilde \bA_m]_{m=1}^M}, \hat{\cW}_u \in \underset{\substack{f \in \calF(L, \bd_{\text{in}}, 1, \bd_{\text{app}}, V, B) \\ \cW_u \in \RR^{\times_{m=1}^M K_m \times_{m=1}^M D_m} }}{\arg\min} \left\{\cR_n(f, \cW_u) + \cP_n(\cW_u;\rho_{\lambda})\right\} \quad \text{up to } \delta_{\text{opt}}, 
\end{equation*}
where $\cR_n, \cP_n$ are defined in \eqref{eqn:TTN-penalty}, and $\rho_{\lambda}=\lambda \psi_{\tau}$ is chosen to be clipped-$L_1$. Assume the depth $L=O(1)$ and the tuning parameters obey $\lambda \ge c_1 \frac{\log Qn + L \log(B(W+1))}{n}$ and $\tau^{-1} \ge c_2(B(W+1))^{L+1} Q n$ for some universal constants $c_1, c_2>0$. Then, for $\forall t>0$, with probability at least $1 - 3e^{-t}$,
\begin{align*}
\EE \big|\hat{f}(\cX)-f^*(\cC, \cV_{\cJ}) \big|^2 &\lesssim \underbrace{W^{-2\beta/d_{\text{eff}}}}_{\text{network approx.}} + \underbrace{\gamma^{-2}\frac{\sum_{m=1}^M \bar{R}_m D_m}{\prod_{m=1}^M D_m}}_{\text{core propagation}} + \underbrace{\Lambda_n + \lambda|\cJ|}_{\text{model/selector complexity}} + \delta_{\text{opt}} +\frac{t}{n},
\end{align*}
where $\Lambda_n \asymp \frac{LW^2\left[ L \log(B(W+1)) + \log n \right]}{n}$, $\beta$ is from Condition \ref{cond:Holder} with the effective dimension $d_{\text{eff}}=\prod_{m=1}^M R_m+|\cJ|$.
\end{theorem}
\noindent Detailed proofs are provided in Section~\ref{sec:theory-tnn} of the supplemental material.

\smallskip
\noindent
\textbf{Error Decomposition and Interpretation.} The bound decomposes the prediction error into three interpretable components:
\begin{itemize}
\item {\em Approximation error.}
The term $W^{-2\beta / d_{\text{eff}}}$ reflects the expressive power of the neural network class in approximating the true regression function. It depends on the effective dimension $d_{\text{eff}}$, which combines the low-rank and sparse components.
\item {\em Core estimation error.}
The term $\mathcal{E}_{\text{core}}$ quantifies the error introduced by estimating the latent core tensor $C$ from the observed tensor $X$. This term captures how uncertainty in the low-rank structure propagates into prediction.
\item {\em Model complexity and selection cost.}
The term $\Lambda_n + \lambda |J|$ reflects the statistical complexity of the neural network and the cost of selecting relevant refinement features. The sparsity penalty controls the size of the active refinement set $J$, thereby regularizing the model.
\end{itemize}

\begin{remark}
Theorem \ref{thm:optimal_inequality} highlights a key advantage of the dual-channel framework: the effective dimension governing the learning rate is
$
d_{\text{eff}} = \prod_{m=1}^M R_m + |J|,
$
which can be substantially smaller than the ambient tensor dimension $\prod_{m=1}^M D_m$. As a result, the DC-TNN estimator avoids the curse of dimensionality by leveraging both low-rank structure and sparse refinement.

Moreover, the decomposition explicitly separates the effects of function approximation, structure estimation, and feature selection, providing a clear understanding of how each component contributes to the overall error. This structure is unique to the dual-channel formulation and is not available in models that rely solely on low-rank or sparse assumptions.
\end{remark}

\subsection{Optimal Rates and Effective Dimension}

We now specialize the bound in Theorem \ref{thm:optimal_inequality} to derive optimal convergence rates and characterize the role of the hybrid structure through the effective dimension.
By appropriately choosing the network width and regularization parameters, the DC-TNN estimator achieves the following rate.

\begin{corollary}
Under the conditions of Theorem \ref{thm:optimal_inequality}, suppose the network width is chosen as $W^* \asymp (\frac{n}{\log n})^{d_{\text{eff}}/(2\beta+d_{\text{eff}})}$ and $\lambda^* \asymp \frac{\log Q + L \log(B(W^* + 1))}{n} \asymp \frac{\log Q + \log n}{n}$, and further assume that the optimization error $\delta_{opt}$ is of smaller order, then we obtain the optimal risk rate:
\begin{align*}
\EE\left|\hat{f}(\cX - f^*(\cC, \cV_{\cJ})\right|^2 \lesssim \left(\frac{\log n}{n}\right)^{\frac{2\beta}{2\beta + d_{\text{eff}}}} + \gamma^{-2} \frac{\sum_{m=1}^M \bar{R}_m D_m}{\prod_{m=1}^M D_m} + \frac{|\mathcal{J}|(\log Q + \log n)}{n}.
\end{align*}
\end{corollary}

\begin{remark}
Theorem~\ref{thm:optimal_inequality} and Corollary~\ref{thm:tight} establish that the DC-TNN estimator achieves the minimax-optimal risk rate for nonparametric regression while explicitly accounting for both low-rank and sparse tensor structures.  
The bound decomposes into three interpretable components:  
(i)~the \emph{network approximation} term reflects the expressive power of the dual-channel network class;  
(ii)~the \emph{core propagation} term quantifies error from estimating the latent low-rank subspace; and  
(iii)~the \emph{model complexity} term captures the statistical cost of sparsity selection in the refinement channel.  
Together, these results show that leveraging the core-refinement architecture yields faster convergence and tighter generalization bounds than conventional tensor neural networks that rely solely on low-rank representations.
\end{remark}

\subsection{Why Dual-Channel Modeling Improves Prediction}

The theoretical results above provide insight into why the proposed dual-channel architecture leads to improved predictive performance in heterogeneous tensor settings. In particular, the error decomposition in Theorem~\ref{thm:optimal_inequality} reveals that the DC-TNN framework effectively balances two competing objectives: capturing global structure and preserving local flexibility.

\smallskip
\noindent
\textbf{Bias-Variance Tradeoff under Structural Modeling.} 
A central challenge in tensor learning is the tradeoff between structural bias and statistical variance: Low-rank models impose strong global structure, reducing dimensionality and variance, but may suffer from representation bias when localized signals are not aligned with the low-rank subspace. High-dimensional or sparse models offer flexibility to capture local effects, but operate in a large ambient space, leading to inflated effective dimension and higher variance.

The dual-channel formulation addresses this tradeoff by decomposing the input into a low-rank core and a sparse refinement component, and modeling them jointly. As reflected in the effective dimension $d_{\text{eff}} = \prod_{m=1}^M R_m + |J|$, the model achieves dimensionality reduction through the core while retaining flexibility through the refinement.

\smallskip
\noindent
\textbf{Error Decomposition Perspective.} The finite-sample risk bound highlights three sources of error: approximation error, governed by the effective dimension $d_{\text{eff}}$;
core estimation error, arising from recovering the latent low-rank structure; and model complexity, driven by the size of the refinement set $J$. This decomposition shows that the dual-channel architecture allows these components to be controlled separately. In contrast, models based solely on low-rank or sparse assumptions entangle these effects, limiting their ability to adapt to heterogeneous data.

\smallskip
\noindent
\textbf{Adaptivity to Heterogeneous Structure.} The DC-TNN framework adapts naturally to different structural regimes: When the data are primarily low-rank, the refinement component becomes negligible, and the model reduces to a low-dimensional representation with strong statistical efficiency; When localized effects are important, the refinement channel selectively incorporates relevant features, avoiding the need to model the entire tensor space. This adaptivity enables the model to achieve near-optimal performance across a broad class of data-generating mechanisms, without requiring prior knowledge of the true structure.

\smallskip
\noindent
\textbf{Comparison with Single-Structure Models.}
The above analysis explains why the dual-channel approach outperforms models that rely on a single structural assumption: Compared to pure low-rank tensor models, DC-TNN avoids bias introduced by unmodeled local effects; Compared to high-dimensional or sparse models, DC-TNN reduces variance by exploiting low-rank structure; Compared to existing tensor neural networks, which typically adopt a single decomposition, DC-TNN provides a strictly richer function class that can represent both global and local patterns.

In summary, the dual-channel architecture improves prediction by combining the strengths of low-rank and sparse modeling while mitigating their respective weaknesses. The theoretical guarantees show that this hybrid structure leads to reduced effective dimension, improved approximation capability, and optimal statistical rates, providing a principled foundation for modeling heterogeneous tensor data.

\section{Structure-Aware Uncertainty Quantification}
\label{sec:uq}

In many applications, predictive performance is evaluated using ROC curves, sensitivity, specificity, and the area under the curve (AUC) \citep{fawcett2006introduction,majnik2013roc}. While point estimates of these quantities are widely used, they do not capture uncertainty arising from finite samples and model estimation, which is critical for reliable decision-making.

Standard approaches to uncertainty quantification for ROC curves include bootstrap methods and parametric modeling assumptions \citep{nakas2023roc}. However, bootstrap-based procedures have been shown to be unreliable in classification settings, and parametric approaches are often misspecified in high-dimensional tensor problems \citep{zheng2025classification}. Recent advances based on conformal prediction provide a distribution-free alternative, but existing methods typically treat tensor inputs as unstructured vectors, ignoring their underlying structure \citep{wu2024conditional,zheng2024quantifying}.

As a result, applying standard conformal methods directly to tensor data often leads to overly conservative and less informative uncertainty estimates, since the calibration step does not exploit the intrinsic geometry of the data.

\subsection{Structure-Aware Conformal Inference}
To address this limitation, we develop a structure-aware conformal inference framework that leverages the core-refinement decomposition of the DC-TNN model. 
We present the method in a binary classification problem with outcome $y \in \{0,1\}$. The framework can be readily extended to a $K$-class classification setting by constructing $(K-1)$ ROC curves, each comparing one category against all others \citep{everson2006multi}.

We randomly split the dataset $\cD = \{(\cX_i, y_i)\}_{i=1}^n$ into training, calibration, and test sets $\cD_{\text{tr}}, \cD_{\text{ca}}, \cD_{\text{tst}}$, with corresponding index sets $\cI_{\text{tr}}, \cI_{\text{ca}}, \cI_{\text{tst}}$ and sizes $n_{\text{tr}} = |\cI_{\text{tr}}|, n_{\text{ca}} = |\cI_{\text{ca}}|$, and $n_{\text{tst}} = |\cI_{\text{tst}}|$. 
A DC-TNN model is trained on $\mathcal{D}_{\text{tr}}$, yielding predicted probabilities $\widehat{\pi}(\mathcal{X}) = \widehat{m}(\mathcal{C}, \cU)$, and construct conformal prediction intervals using a split conformal framework.
We aim to construct \emph{conditional prediction ROC (CP-ROC) bands} that quantify uncertainty in sensitivity (True Positive Rate) and specificity (True Negative Rate) across decision thresholds $\lambda$, defined as follows
\[
{\rm Sens}(\lambda) 
= \frac{1}{|\mathcal{I}^{(1)}_{tst}|}\sum_{j\in\mathcal{I}^{(1)}_{tst}}
\mathbf{1}\!\big(\hat\pi(\cX_j)>\lambda\big),\qquad
{\rm Spec}(\lambda) 
= \frac{1}{|\mathcal{I}^{(0)}_{tst}|}\sum_{j\in\mathcal{I}^{(0)}_{tst}}
\mathbf{1}\!\big(\hat\pi(\cX_j)\le\lambda\big).
\]
For simplicity, we write $\hat{\pi}_i = \hat{\pi}(\cX_i)$ when no ambiguity arises.

\smallskip
\noindent
\textbf{Step 1: Local calibration in core-refinement space.}
For each test point $j\in\mathcal{I}_{tst}$, construct a local calibration set $\mathcal{I}^{loc}_{ca,j}\subset\mathcal{I}_{ca}$ using $K$-nearest neighbors under a distance
\begin{equation}\label{eqn:fa-distance}
d_{CR}(i,j)=\|\hat\cC(\cX_i)-\hat\cC(\cX_j)\|_F
+ \omega\|\hat\cU(\cX_i)-\hat\cU(\cX_j)\|_F,
\end{equation}
with given $\omega>0$. This metric adapts to the two-channel decomposition: low-rank similarity dominates when the signal is factor-driven, while selected idiosyncratic modes refine local neighborhoods. 

\smallskip
\noindent
\textbf{Step 2: Class-conditional calibration.}
The ideal conformity score is defined as $s_i = \pi(\cX_i) - \hat{\pi}_i$, 
where $\pi(\cX_i) = \mathbb{P}(y=1 \mid \cX_i)$ denotes the true conditional probability. 
Since $\pi(\cX_i)$ is not observable, we approximate it using a nonparametric estimator constructed in the core-refinement latent space.

Specifically, we define $\tilde{\pi}(\cdot)$ as a $K_{\mathrm{tr}}$-nearest neighbor (KNN) smoother based on $\hat{\pi}$ over the training set $\cD_{\mathrm{tr}}$, where neighbors are selected according to the distance in \eqref{eqn:fa-distance}. This yields a proxy $\tilde{\pi}(\cX_i) = \tilde{\pi}\big(\hat {\cC}(\cX_i),\hat{\cU}(\cX_i)\big)$ for $\pi(\cX_i)$. The choice of $K_{\mathrm{tr}}$ is discussed in Remark~\ref{remark:knn_parameters}.

For each $i \in \mathcal{I}^{\mathrm{loc}}_{ca,j}$, we define the estimated conformity score $\tilde{s}_i = \tilde{\pi}\big(\hat {\cC}(\cX_i),\hat{\cU}(\cX_i)\big) - \hat{\pi}_i$,
and construct the class-conditional calibration sets
\[
\mathcal{S}^{(1)}_j = \{ \tilde{s}_i : i \in \mathcal{I}^{\mathrm{loc}}_{ca,j},\, y_i = 1 \}, 
\qquad
\mathcal{S}^{(0)}_j = \{ \tilde{s}_i : i \in \mathcal{I}^{\mathrm{loc}}_{ca,j},\, y_i = 0 \}.
\]

\smallskip
\noindent
\textbf{Step 3: Conformal intervals for predicted probabilities.} 
Let $q^{(k)}_{\gamma}(j)$ be the $\gamma$-quantile of $\mathcal{S}^{(k)}_j$.  
The class-conditional conformal interval for $\pi(\cX_j)$ is
\[
C_{\alpha}^{(k)}(\cX_j) 
=\big[\hat\pi(\cX_j)+q^{(k)}_{\alpha/2}(j),\;
\hat\pi(\cX_j)+q^{(k)}_{1-\alpha/2}(j)\big]\cap[0,1],\quad k\in\{0,1\}.
\]

\smallskip
\noindent
\textbf{Step 4: Pointwise ROC confidence bands.} 
We then combine all the individual intervals into the ROC confidence band and form the AUC confidence interval. Let $\mathcal{I}^{(1)}_{tst}=\{j\in\mathcal{I}_{tst}:y_j=1\}$ and $\mathcal{I}^{(0)}_{tst}=\{j\in\mathcal{I}_{tst}:y_j=0\}$, denote $ \underline{C},\;\overline{C}$ as the lower and upper limit of an interval $C$.
For threshold $\lambda\in[0,1]$, obtain the confidence intervals for sensitivity and specificity by
\[
C^{\mathrm{sens}}_{\alpha}(\lambda)=\left[\frac{1}{|\mathcal{I}^{(1)}_{tst}|}\sum_{j\in\mathcal{I}^{(1)}_{tst}}
\mathbf{1}\{\underline{C_{\alpha}^{(1)}}(\cX_j)>\lambda\},\;\;
\frac{1}{|\mathcal{I}^{(1)}_{tst}|}\sum_{j\in\mathcal{I}^{(1)}_{tst}}
\mathbf{1}\{\overline{C_{\alpha}^{(1)}}(\cX_j)>\lambda\}\right],
\]
\[
C^{\mathrm{spec}}_{\alpha}(\lambda)=\left[\frac{1}{|\mathcal{I}^{(0)}_{tst}|}\sum_{j\in\mathcal{I}^{(0)}_{tst}}
\mathbf{1}\{\overline{C_{\alpha}^{(0)}}(\cX_j)\le\lambda\},\;\;
\frac{1}{|\mathcal{I}^{(0)}_{tst}|}\sum_{j\in\mathcal{I}^{(0)}_{tst}}
\mathbf{1}\{\underline{C_{\alpha}^{(0)}}(\cX_j)\le\lambda\}\right].
\]
Varying $\lambda$ over a grid for $C_{\alpha}^{sens}(\lambda)$ yields ROC confidence band for sensitivity and for $C_{\alpha}^{spec}(\lambda)$ yields ROC confidence band for specificity. 

\smallskip
\noindent
\textbf{Step 5: AUC confidence intervals. }
For AUC, the two types of confidence intervals are calculated from the numerical integration of the upper and lower bounds of the ROC confidence bands, for either the sensitivity or the specificity. 
\[
C_{\rm auc}^{\rm sens}(\alpha) = \left[\int_0^1 
\underline{C_{\alpha}^{sens}}\big({\rm Spec}^{-1}(1-x)\big)dx,\;\; \int_0^1 
\overline{C_{\alpha}^{sens}}\big({\rm Spec}^{-1}(1-x)\big)dx\right]
\]
\[
C_{\rm auc}^{\rm spec}(\alpha) = \left[1-\int_0^1 
\overline{C_{\alpha}^{spec}}\big({\rm Sens}^{-1}(\cX)\big)dx,\;\; 1-\int_0^1 
\underline{C_{\alpha}^{spec}}\big( {\rm Sens}^{-1}(\cX)\big)dx\right]
\]

\subsection{Theoretical Coverage Guarantees}

We now establish the validity of the proposed structure-aware conformal procedure.
We first define the oracle sensitivity and specificity as our target: 
\[
{\rm Sens}_0(\lambda) 
= \frac{1}{|\mathcal{I}^{(1)}_{tst}|}\sum_{j\in\mathcal{I}^{(1)}_{tst}}
\mathbf{1}\!\big(\pi(\cX_j)>\lambda\big),\qquad
{\rm Spec}_0(\lambda) 
= \frac{1}{|\mathcal{I}^{(0)}_{tst}|}\sum_{j\in\mathcal{I}^{(0)}_{tst}}
\mathbf{1}\!\big(\pi(\cX_j)\le\lambda\big),
\]
where $\pi(\cX)=P(y=1\mid \cX)$ is the oracle success probability.

\begin{proposition}[Coverage of structure-aware ROC bands]\label{prop: cov CP_ROC} 
Suppose for each $j\in\mathcal{I}_{tst}$ the local calibration set $\mathcal{I}^{loc}_{ca,j}$ is exchangeable with $(\cX_j,y_j)$ and 
$\lim_{n_{tr}\to\infty}\max_{j \in I_{\mathrm{tst}}}|\tilde{\pi}\big(\hat {\cC}(\cX_j),\hat{\cU}(\cX_j)\big) - \pi(\cX_j)|=0$. 
Then for randomly chosen $\lambda\in[0,1]$,
\begin{align*}    &\lim_{n_{tr},n_{ca},n_{tst}\to\infty}\min\left\{P\!\big({\rm Sens}_0(\lambda)\in C^{\mathrm{sens}}_{\alpha}(\lambda)\big),\;\;P\!\big({\rm Spec}_0(\lambda)\in C^{\mathrm{spec}}_{\alpha}(\lambda)\big)\right\}\ge 1-2\alpha
\end{align*}
Additionally, if we further assume that the CDF of $s_i$ with $y_i=k$, denoted as $F_k(\cdot)$, satisfies that $ F_{k}^{-1}(\alpha/2) < 0 $ and $ F_{k}^{-1}(1 - \alpha/2) > 0$, then as $n_{tr},n_{ca},n_{tst}\to\infty$ almost surely, ${\rm Sens}(\lambda) \in C^{\mathrm{sens}}_{\alpha}(\lambda)$ and $ {\rm Spec}(\lambda) \in C^{\mathrm{spec}}_{\alpha}(\lambda)$.
\end{proposition}

\begin{remark}[\bf Exchangeability under local calibration]
The assumption that the local calibration set $\mathcal{I}^{\mathrm{loc}}_{\mathrm{ca},j}$ is exchangeable with the target point $(\cX_j, y_j)$ assumes that the nearest neighbors form a locally homogeneous sample. This condition is supported by the nearest-neighbor conformal prediction framework of \cite{gyorfi2019nearest}, where conditional coverage is achieved under a smooth conditional distribution of $y_j$ given $\cX_j$ and an appropriately chosen k-NN algorithm.
\end{remark}

\begin{remark}[\bf Consistency of probability estimation]\label{remark:concistency_prob_est}
The assumption in Proposition~\ref{prop: cov CP_ROC} requires that the estimated probability
$\tilde{\pi}(\hat{\cC}(\cX), \hat{\cU}(\cX))$ approximates the true
probability $\pi(\cX)$ uniformly over the test sample. This is stronger
than classical $L_1$-consistency of $K$-nearest neighbor regression, but is substantially weaker than requiring uniform convergence over the entire input space.

This requirement is motivated by performing probability smoothing in the core-refinement representation space rather than in the ambient tensor space. Once $\cX$ is mapped to $(\hat{\cC}(\cX), \hat{\cU}(\cX))$, the
effective dimension is significantly reduced, making nonparametric
approximation more stable. Under mild smoothness conditions on $\pi(\cX)$
as a function of this representation, consistency over the test sample
becomes plausible. Remark~7 further clarifies this connection through
standard nonparametric convergence rates.
\end{remark}

\begin{remark}[\bf Choice of KNN parameters and connection to nonparametric rates] \label{remark:knn_parameters}
Under additional smoothness assumptions, the consistency requirement in Proposition~\ref{prop: cov CP_ROC} can be related to standard nonparametric regression theory in the core-refinement representation space. Suppose $\pi(\cX)$ is $(\beta, C)$-H\"older continuous with respect to the metric induced by $(\hat{\cC}(\cX), \hat{\cU}(\cX))$, and that the representation has effective dimension $d_{\text{rep}} = \prod_{m=1}^M \bar R_m + \prod_{m=1}^M \bar K_m$. 
Then the $K_{\mathrm{tr}}$-nearest neighbor estimator satisfies
\[
\sup_{j \in I_{\mathrm{tst}}}
\left|
\tilde{\pi}(\hat{\cC}(\cX_j), \hat{\cU}(\cX_j))
-
\pi(\cX_j)
\right|
=
O_p\!\left(
\left(\frac{K_{\mathrm{tr}}}{n_{\mathrm{tr}}}\right)^{\beta/d_{\mathrm{rep}}}
+
K_{\mathrm{tr}}^{-1/2}
\right),
\]
up to logarithmic factors and standard regularity conditions.

Balancing the bias and variance terms yields
\[
K_{\mathrm{tr}} \asymp n_{\mathrm{tr}}^{\frac{2\beta}{2\beta + d_{\mathrm{rep}}}},
\qquad
\max_{j \in I_{\mathrm{tst}}}
\left|
\tilde{\pi}(\hat{\cC}(\cX_j), \hat{\cU}(\cX_j))
-
\pi(\cX_j)
\right|
=
O_p\!\left(n_{\mathrm{tr}}^{-\beta/(2\beta + d_{\mathrm{rep}})}\right).
\]

Thus, the assumption in Proposition~\ref{prop: cov CP_ROC} is most plausible when the learned representation is sufficiently low-dimensional and preserves the smoothness structure of $\pi(\cX)$. In the special case $\beta = 1$, this reduces to
$K_{\mathrm{tr}} \asymp n_{\mathrm{tr}}^{2/(2 + d_{\mathrm{rep}})}$.

For local calibration, \cite{gyorfi2019nearest} suggests that $K_{\text{tr}}/\big(\log(n_{tr})\big)^2\to\infty$ and $K_{\text{tr}}\big(\log(n_{tr})\big)^{d_{rep}}/n\to 0$ to ensure consistent estimation of class-conditional quantiles. 
In practice, $K_{\mathrm{tr}}$ and $K_{\mathrm{ca}}$ are chosen to balance approximation accuracy and sufficient local sample size.
\end{remark}

For the coverage rate of the AUC confidence intervals, similarly define the oracle AUC, denoted as ${\rm AUC}_0$, as the area under the oracle ROC curve defined by ${\rm Sens}_0(\cdot)$ and ${\rm Spec}_0(\cdot)$. And the theoretical coverage guarantee is given below.
\begin{corollary}[AUC coverage from CP-ROC bands]\label{cor: AUC confidence band}
	Assume the same conditions as in Proposition~\ref{prop: cov CP_ROC} hold, then
	\begin{align*}    &\lim_{n_{tr},n_{ca},n_{tst}\to\infty}\min\left\{P\big({\rm AUC}_0\in C^{\mathrm{sens}}_{\rm auc}(\alpha)\big),\;\;P\big({\rm AUC_0}\in C^{\mathrm{spec}}_{\rm auc}(\alpha)\big)\right\}\ge 1-2\alpha
	\end{align*}  
\end{corollary}

The proofs are detailed in Section \ref{sec:theory-uq} in the supplemental material. 
The proposed uncertainty quantification framework leverages the same core-refinement structure used for prediction to construct {\em valid and informative confidence regions}. By calibrating in a structured feature space, the method improves upon standard conformal approaches and provides a principled way to quantify uncertainty in tensor-based predictive models.

\section{Provable Low-Rank Structure Selection} \label{sec:selector}

Identifying the true low-rank structure of tensor features is a fundamental yet unresolved challenge:~naive selection based on predictive accuracy or AUC lacks statistical validity and often leads to overfitting.  
We propose a \emph{conformal ROC method} that leverages the DC-TNN’s core-refinement representation to construct finite-sample confidence bands for performance differences between candidate models.  
This procedure provides the first provably valid statistical test for selecting among Tucker, CP, or other low-rank tensor structures.

To illustrate the idea, we compare two fitted models, $\hat f^{(\mathrm{Tucker})}$ and $\hat f^{(\mathrm{CP})}$, trained on the same dataset. Specifically, we consider the following hypothesis test problem:
\begin{align*}
    H_0: \hat f^{(CP)}\text{ outperforms } \hat f^{(Tucker)}, H_a: \hat f^{(CP)} \text{ no better than } \hat f^{(Tucker)}.
\end{align*}
Notice that the above hypothesis test protects the Null, meaning that we prefer $\hat f^{(Tucker)}$ if there is no evidence that $\hat f^{(CP)}$ is significantly better. On the other hand, if $\hat f^{(CP)}$ is preferred, the symmetric hypothesis test can be adopted where $H_{0}: \hat f^{(Tucker)}\text{ outperforms } \hat f^{(CP)}$.
We focus on binary classification with $y\in\{0,1\}$, noting that the approach generalizes to multi-class problems by constructing $K$ ROC curves, one for each label $k=1,\dots,K$.  

\subsection{Conformal ROC-Based Structure Selection for Classification}

The dataset is partitioned into training, calibration, and test subsets, 
$\mathcal D_{tr}\cup\mathcal D_{ca}\cup\mathcal D_{tst}$.  
Models are trained on $\mathcal D_{tr}$ to obtain 
$\hat f^{(\mathrm{Tucker})}(\cdot\mid\mathcal D_{tr})$ and 
$\hat f^{(\mathrm{CP})}(\cdot\mid\mathcal D_{tr})$, 
and the calibration set $\mathcal D_{ca}$ is used to construct conformal confidence bands that support inference on the held-out test set $\mathcal D_{tst}$.
For a fixed threshold $\lambda'\in[0,1]$, the two models yield classifiers $\hat y^{(Tucker)} = \bbone\{\hat f^{(Tucker)}(\cX)\ge \lambda'\}$ and $\hat y^{(CP)} = \bbone\{\hat f^{(CP)}(\cX)\ge \lambda'\}$.
A naive comparison would be based on the misclassification rate
$\frac{1}{n_{tst}} \sum_{j\in\mathcal I_{tst}} 1(y_j \neq \hat y)$, 
but this rate neglects the potential effects arising from the choice of threshold $\lambda'$. By varying $\lambda'$, one obtains the ROC curves of both models, but direct statistical tests for comparing these curves under a fitted classifier setting are lacking. 
Existing methods for comparing ROC curves \citep{delong1988comparing,venkatraman1996distribution} are tailored to diagnostic testing on fixed scores and do not account for the uncertainty introduced by model training.

To address this, we consider the difference in predictive scores
\begin{equation*}
d(\cX) = \hat f^{(Tucker)}(\hat\cC^{(Tucker)},\hat\cU^{(Tucker)}) - \hat f^{(CP)}(\hat\cC^{(CP)},\hat\cU^{(CP)}).
\end{equation*}
For each test point, $d(\cX)$ induces the contingency table Table \ref{tab: contingency tab}, from which a ROC curve is constructed. 
To interpret this contingency table, if Tucker decomposition truly outperforms CP, then at any threshold $\lambda$ we expect more positives to satisfy $d(\cX)>\lambda$ and more negatives to satisfy $d(\cX)\le \lambda$. In terms of the contingency table, this corresponds to large $n_{11}(\lambda)$ and $n_{22}(\lambda)$, with comparatively small $n_{12}(\lambda)$ and $n_{21}(\lambda)$. As $\lambda$ varies, the resulting ROC curve based on $d(\cX)$ will lie above the reference line $y=x$, indicating that Tucker systematically assigns higher scores to the true positives relative to CP. Conversely, if CP is superior, then $d(\cX)$ tends to be smaller for positives and larger for negatives, leading to an ROC curve that falls below the diagonal. When the two decompositions have comparable predictive ability, the difference curve will fluctuate around the diagonal, with no systematic dominance in sensitivity–specificity trade-offs.

\begin{table}[ht]
\centering
\begin{tabular}{c|cc}
& $d(\cX)>\lambda$ & $d(\cX)\le\lambda$  \\
\hline
$y=1$& $n_{11}(\lambda)$ &$n_{12}(\lambda)$ \\ 
$y=0$& $n_{21}(\lambda)$ &$n_{22}(\lambda)$\\ 
\end{tabular}
\caption{Contingency table under our model comparison setup}
\label{tab: contingency tab}
\end{table}

To rigorously account for finite-sample variability, we can construct conformal confidence bands for the difference ROC curve with similar procedure as described in Section \ref{sec:uq}. These bands provide a formal guarantee: with probability at least $1-\alpha$, the oracle difference ROC curve is contained entirely within the band. If the band lies strictly above (or below) the diagonal, one can conclude that Tucker (or CP) is significantly better; if the band overlaps the diagonal, the two models are statistically indistinguishable. The procedure is summarized in Algorithm~\ref{algo:Selector}. A key component is the use of local calibration: for each test point $\cX_j$, we form a neighborhood $\mathcal N_j = {(\cX_i,y_i)\in\mathcal D_{ca}: d_{FA}(\cX_i,\cX_j)\le \delta,\ y_i=y_j}$ based on a distance metric $d_{FA}(\cdot,\cdot)$ \eqref{eqn:fa-distance} in the core-refinement-augmented latent space. This ensures approximate exchangeability between calibration and test points, which is necessary for conformal validity of the ROC band.

\begin{algorithm}[htbp!]
\caption{Model Selector}\label{algo:Selector}
\DontPrintSemicolon
  \KwInput{Training set $\{(\bcalX_i,y_i):i\in\mathcal{I}_{ tr}\}$, calibration set $\{(\bcalX_i,y_i):i\in\mathcal{I}_{ca}\}$ and test set $\{(\bcalX_j,y_j):i\in\mathcal{I}_{tst}\}$, split its index set by the label y: $\mathcal{I}_{tst,0}$ and $\mathcal{I}_{tst,1}$, significance level \(\alpha\in(0,1)\), output difference between the two models $d(\bcalX)$, $k$-NN selector $\mathcal{N}(\bcalX)$, grid value $-1<\lambda_1< \lambda_2<\cdots < \lambda_G\le 1$}
  \KwOutput{Testing ROC curve band and its AUC confidence band.}
  \For{$j\in\mathcal{I}_{tst}$}{
  Calculate the local calibration set $\mathcal{N}(\bcalX_j)\subset \mathcal{D}_{ca}$ and denote $\mathcal{I}_{jk} = \{i: (\cX_i, y_i)\in\mathcal{N}(\cX_j),y_i=k\}$\; 
  Calculate difference $d_i = d(\bcalX_i)$, for $i\in\mathcal{I}_{jk}$ and $d_j = d(\bcalX_j)$\;
  Obtain conformal prediction interval for the difference$d(\bcalX_j)$ with $y_j = k$ as $dC_{\alpha}^{(k)}(\bcalX_j) = [q_{\alpha/2}(\{d_i\}_{i\in\mathcal{I}_{jk}}),q_{1-\alpha/2}(\{d_i\}_{i\in\mathcal{I}_{jk}})]$.}
  \For{$g=1,2,\cdots,G$}{
  Calculate sensitivity and specificity for the difference as 
  \begin{align*}
      {\rm dSens}(\lambda_g) = \frac{1}{|\mathcal{I}_{tst,1}|}\sum_{j\in\mathcal{I}_{tst,1}} {\bf 1}\big(d_j>\lambda_g\big),\quad {\rm dSpec}(\lambda_g) = \frac{1}{|\mathcal{I}_{tst,0}|}\sum_{j\in\mathcal{I}_{tst,0}} {\bf 1}\big(d_j\le \lambda_g\big).
  \end{align*}
  Calculate confidence intervals for sensitivity $Sens(\lambda_g)$  and specificity $Spec(\lambda_g)$ as
  \begin{align*}
    &dC^{\rm sens}_{\alpha}(\lambda_g) = \bigg[\frac{1}{|\mathcal{I}_{tst,1}|}\sum_{j\in\mathcal{I}_{tst,1}} {\bf 1}\big(\underline{dC_{\alpha}^{(1)}}(\cX_j)>\lambda_g\big),\quad \frac{1}{|\mathcal{I}_{tst,1}|}\sum_{j\in\mathcal{I}_{tst,1}} {\bf 1}\big(\overline{dC_{\alpha}^{(1)}}(\cX_j)>\lambda_g\big) \bigg],\\
    &dC^{\rm spec}_{\alpha}(\lambda_g) = \bigg[\frac{1}{|\mathcal{I}_{tst,0}|}\sum_{j\in\mathcal{I}_{tst,0}} {\bf 1}\big(\overline{dC_{\alpha}^{(0)}}(\cX_j)\le \lambda_g\big),\quad \frac{1}{| \mathcal{I}_{tst,0}}| \sum_{j\in \mathcal{I}_{tst,0}} {\bf 1}\big(\underline{dC_{\alpha}^{(0)}}(\cX_j)\le \lambda_g\big) \bigg].
\end{align*}}
Calculate AUC confidence set as
\begin{align*}
    &\resizebox{\textwidth}{!}{$dC_{\rm auc}^{\rm sens}(\alpha) = \bigg[ \sum_{g=2}^G \underline{dC_{\alpha}^{sens}}(\lambda_{g-1}) \big[{\rm dSpec}(\lambda_g)-{\rm dSpec}(\lambda_{g-1})\big], \sum_{g=2}^G \overline{dC_{\alpha}^{sens}}(\lambda_{g-1}) \big[{\rm dSpec}(\lambda_g)-{\rm dSpec}(\lambda_{g-1})\big]\bigg]$}\\
    &\resizebox{\textwidth}{!}{$dC_{\rm auc}^{\rm spec}(\alpha) = \bigg[\sum_{g=2}^G \underline{dC_{\alpha}^{spec}}(\lambda_{g}) \big[{\rm dSens}(\lambda_{g-1})-{\rm dSens}(\lambda_{g})\big], \sum_{g=2}^G \overline{dC_{\alpha}^{spec}}(\lambda_{g}) \big[{\rm dSens}(\lambda_{g-1})-{\rm dSens}(\lambda_{g})\big]\bigg]$}
\end{align*}
\end{algorithm}

\subsection{Finite-Sample Guarantees for Structure Selection}
\label{subsec:selector}

We now establish the statistical validity of the proposed structure selector. First, we show that the conformal confidence intervals for the AUC achieve finite-sample coverage at the target confidence level. Second, we prove that, under mild regularity conditions, the selector is asymptotically consistent: if one of the low rank structure identifies the true structure of the data, then the selector will be able to select it.

Let $f^{(T)*}$ and $f^{(C)*}$ denote the population risk minimizers within the Tucker and CP classes, respectively, and write the \emph{oracle difference score} as
\[
\delta(\cX)=f^{(T)*}(\cX)-f^{(C)*}(\cX).
\]
For $\lambda\in\mathbb R$, define the \emph{oracle difference sensitivity and specificity}:
\[
{\rm dSens}_0(\lambda)=\frac{1}{|\mathcal I_{tst}^{(1)}|}\sum_{j\in\mathcal I_{tst}^{(1)}}\mathbf 1\{\delta(\cX_j)>\lambda\},\qquad
{\rm dSpec}_0(\lambda)=\frac{1}{|\mathcal I_{tst}^{(0)}|}\sum_{j\in\mathcal I_{tst}^{(0)}}\mathbf 1\{\delta(\cX_j)\le\lambda\}.
\]
and the oracle AUC can be calculated by
\[
{\rm dAUC}_0 = \int_0^1 {\rm dSens_0}\big(({\rm dSpec}_0)^{-1}(1-x)\big)dx
\]
Then we develop the theoretical coverage for the AUC confidence intervals as follows:

\begin{theorem}[Validity of different AUC Conformal confidence intervals]\label{thm:diff-roc-fs}

Assume for each $j\in\mathcal I_{tst}^{(k)}$: (i) \emph{local exchangeability}—conditionally on $\cX_j$ and $y_j=k$, the multiset $\{(\cX_i,y_i): i\in\mathcal I^{loc}_{ca,j},\,y_i=k\}$ is exchangeable with $(\cX_j,y_j)$; (ii) the conditional CDF of $s=\delta(\cX)-d(\cX)$ given $(\cX,y=k)$ is continuous. 
Then we have that
  \begin{align*}    &\lim_{n_{ca},n_{tst}\to\infty}\min\left\{P\big({\rm dAUC}_0\in dC^{\mathrm{sens}}_{\rm auc}\big),\;\;P\big({\rm dAUC}_0\in dC^{\mathrm{spec}}_{\rm auc}\big)\right\}\ge 1-2\alpha
\end{align*}  
\end{theorem}
The proof follows a similar argument to that of Corollary~\ref{cor: AUC confidence band}, with full details provided in Section~\ref{sec:theory-selector} of the supplemental material. Note that compare with Corollary~\ref{cor: AUC confidence band}, we do not need a k-NN estimator of $\tilde\pi(\cdot)$ as we directly target on the difference $d(\cX)$. Consequently, the assumptions here are weaker.


Now we are ready to choose the decision rule that define the structure selector. Notice that if Tucker structure has better predictive performance, the difference ROC curve should lie above the reference line $y=x$, thus we have that ${\rm dAUC_0}>0.5$. We can then use the confidence interval $dC_{auc}^{sens}$ to test the predictive performance of the two structures for samples with label $y=1$, and use $dC_{auc}^{spec}$ to test the predictive performance for samples with label $y=0$. The detailed selector is given below:
\begin{itemize}
    \item Sensitivity selector: Focus on sensitivity for positively labeled group, select Tucker if the lower bound of $C_{auc}^{sens}$ is above $0.5$; select CP if the upper bound of $C_{auc}^{sens}$ is below $0.5$; otherwise, declare a tie.
    \item Specificity selector: Focus on specificity for negatively labeled group, select Tucker if the lower bound of $C_{auc}^{spec}$ is above $0.5$; select CP if the upper bound of $C_{auc}^{spec}$ is below $0.5$; otherwise, declare a tie.
\end{itemize}

To develop theoretical consistency of the selector, assume the true low rank structure is either Tucker or CP, then intuitively, the other low rank structure won't be able to provide better prediction than this true structure, thus our proposed selector is able to select the true structure with probability at the chosen confidence level $1-\alpha$. The detailed theorem is given below, and its proof is provided in Section \ref{sec:theory-selector} in the supplemental material. 

\begin{theorem}(False discovery rate control)\label{thm:selector-consistency}\label{thm: selector consistency} 
For $i \in \mathcal{I}_{\mathrm{tst}}$, let $p^*({\mathcal{C}}_i, \mathcal{U}_{i}) = \mathbb{E}[y_i=1\mid \mathcal{C}_i, \mathcal{U}_{i}]$. Assume the Tucker-based predictor $\hat{f}^{\mathrm{(Tucker)}}(\hat{\mathcal{C}}_i^{\mathrm{(Tucker)}}, \hat{\mathcal{U}}_{i}^{\mathrm{(Tucker)}})$ is consistent in the sense that
\begin{equation}\label{eq: (selector) Tucker consistent}
    \lim_{n_{tr}\to\infty}1|\hat f^{(Tucker)}(\hat\cC_i^{(Tucker)},\hat\cU_{i}^{(Tucker)})-p^*(\boldsymbol{\cC_i}, \cU_{i})| = 0.
\end{equation}
If~(\ref{eq: (selector) Tucker consistent}) holds for all $y_i=1$, then
\begin{equation*}
    \lim_{n_{tr},n_{ca},n_{tst}\to\infty}P\left(\text{Sensitivity selector select Tucker or declare a tie}\right)\ge 1-2\alpha
\end{equation*}
If~(\ref{eq: (selector) Tucker consistent}) holds for all $y_i=0$, then
\begin{equation*}
    \lim_{n_{tr},n_{ca},n_{tst}\to\infty}P\left(\text{Specificity selector select Tucker or declare a tie}\right)\ge 1-2\alpha
\end{equation*}
Analogously, if the CP predictor $\hat{f}^{\mathrm{(CP)}}(\hat{\mathcal{C}}_i^{\mathrm{(CP)}}, \hat{\mathcal{U}}_{i}^{\mathrm{(CP)}})$ is consistent (i.e., satisfies the same $o_p(1)$ convergence), then the same conclusions hold for the sensitivity and specificity selectors applied to the CP structure.
\end{theorem}

\section{Numerical Experiments}
\label{sec:simulation}
\noindent\underline{\textbf{Data generation.}} The data generation process (DGP) follows the core-refinement formulation in \eqref{eq:core_refine} and the DC-TNN architecture in \eqref{eqn: dual-channel TNN}, specialized to a tensor covariate $\cX \in \RR^{D_1 \times D_2 \times D_3}$ with a binary response $y \in \{0, 1\}$. In all experiments we set $(D_1, D_2, D_3) = (32, 32, 32)$ and generate $n = 2000$ observations with $n_0 = 1000$ samples labeled $0$ and $n_1 = 1000$ samples labeled $1$. 

Each tensor is generated from the decomposition $\cX = \cS(\cC) + \cU$, where the low-rank signal $\cS(\cC)$ is determined by either the Tucker reconstruction map in \eqref{eqn:tnn-tucker} or the CP reconstruction map in \eqref{eqn:cp decomposition}. We consider two ground-truth regimes. In the Tucker regime, the signal is $\cS(\cC) = \cC \times_1 \bU_1 \times_2 \bU_2 \times_3 \bU_3$, where $\bU_m \in \RR^{32 \times 3}$ has orthonormal columns obtained by QR orthonormalization of an i.i.d. uniform matrix with entries in $[-5, 5]$; and the Tucker core $\cC \in \RR^{3 \times 3 \times 3}$ is generated by 
first sampling i.i.d. standard Gaussian entries, applying Gaussian kernel smoothing with bandwidth $\sigma = 1,0$, and then rescaling to achieve Frobenius norm $\norm{\cC} = 5.0$. In the CP regime, the signal is $\cS(\cC) = \sum_{r=1}^R c_r \ba_{r1} \circ \ba_{r2} \circ \ba_{r3}$ with CP rank $R = 12$, where each factor matrix $\bA_m = [\ba_{m1}, \ldots, \ba_{mR}]$ is initialized with orthonormal columns $\{\tilde \ba_{rm}\}$ via QR and then perturbed to be non-orthogonal: $\ba_{1m} = \tilde \ba_{1m}$ and $\ba_{rm} = (\tilde \ba_{1m} + \eta \tilde \ba_{rm})/ \norm{\ba_{1m} + \eta \ba_{rm}}_2$ for $r \ge 2$, where $\eta = (\vartheta^{-2/3} - 1)^{1/2}$, $\vartheta = \delta / (r-1)$ and $\delta=0.1$. The CP coefficients $\bc = (c_1, \ldots, c_R)$ are generated from an AR(1)-type recursion: $c_1 \sim \mathcal{N}(0, 1)$ and $c_r = \rho c_{r-1} + \sqrt{1 - \rho^2} \varepsilon_r$ for $r \geq 2$ where $\varepsilon_r \stackrel{\text{i.i.d.}}{\sim} \cN(0, 1)$ and $\rho = 0.7$, followed by rescaling to achieve $||\bc||_2 = 8.0$; the CP ``core'' is represented as a super-diagonal tensor $\cC \in \RR^{R \times R \times R}$. The residual $\cU$ is decomposed as $\cU = \cU_S + \cU_N$ into a sparse refinement component $\cU_S$ and a noise (nuisance component) $\cU_N$. We first sample a fixed active set $J \subset [32] \times [32] \times [32]$ uniformly at random without replacement with $|J| = 18$. Conditional on $J$, we generate the nuisance term by sampling i.i.d. from $\cN(0, \sigma_N^2)$ with $\sigma_N = 0.1$. The sparse refinement $\cU_S$ is supported on $J$ and is constructed as follows: let $j_1,\dots,j_{18}$ denote the $18$ elements of $J$. For each $i=1,\dots,18$, we draw an independent sign $\xi_i \in \{+1, -1\}$, a scale $a_i \sim \text{Unif}[5.0, 8.0]$, and set $\cU_S(j_i) = \xi_i a_i |\cS(j_i)|$; all inactive coordinates are set to zero.  

Finally, labels are generated by a ground-truth DC-TNN mapping $\pi(\cX) = \sigma(z(\cC, \cU_D))$, where $\sigma$ is sigmoid function, $z(\cC, \cU_D)$ is the scalar logit produced by a two-hidden-layer ($L = 2$) dual-channel tensor network with ReLU activations. To align with the refinement-channel input $\cU_D$, we additionally construct an oracle dense refinement tensor $\cU_D \in \mathbb{R}^{2 \times 3 \times 3}$ by first extraction nonzero entries of $\cU_S$ and then reshape to $\cU_D \in \RR^{2\times 3\times 3}$ such that $2 \cdot 3 \cdot 3 = |J|$; and the ordering of indices in $J$ is fixed once at the beginning of data generation. The class-balanced dataset is obtained by repeated candidate generation and acceptance into class buckets until both buckets are filled. The following table summarizes key statistics from the data generation process for both Tucker and CP regimes:

\begin{table}[h]
\centering
\caption{Data Generation Summary Statistics}
\label{tab:dgp_summary}
\begin{tabular}{lcccc}
\hline
Structure & $\bar{z}|y{=}1$ & $\bar{z}|y{=}0$ & $\bar{\pi}|y{=}1$ & $\bar{\pi}|y{=}0$ \\
\hline
Tucker-generated dataset & $2.542_{(2.176)}$ & $-2.256_{(1.938)}$ & $0.825_{(0.235)}$ & $0.189_{(0.256)}$ \\
CP-generated dataset & $3.048_{(2.810)}$ & $-2.724_{(2.351)}$ & $0.820_{(0.244)}$ & $0.176_{(0.225)}$ \\
\hline
\multicolumn{5}{l}{\footnotesize Note: Values shown as mean$_{(\text{std})}$}
\end{tabular}
\end{table}

\noindent\underline{\textbf{Estimation.}} For each generated dataset, we fit the core-refinement DC-TNN estimator by the two-stage procedure described in Section~\ref{subsec:training}, implementing either Tucker or CP decomposition to estimate the core tensors. This yields four experimental regimes: we fit Tucker-core and CP-core DC-TNNs on the Tucker-generated dataset, and we fit Tucker-core and CP-core DC-TNNs on the CP-generated dataset. 

For each dataset, we split the $n = 2000$ observations into a training set, a calibration set, and a test set with fractions 0.6/0.2/0.2, yielding $n_{\text{tr}} = 1200$, $n_{\text{ca}} = 400$, and $n_{\text{tst}} = 400$. All tensor inputs are centered by the empirical mean of the training tensors, i.e., $\widetilde{\mathcal{X}}_i = \mathcal{X}_i - \overline{\mathcal{X}}_{\text{tr}}$. In the Tucker-core estimation, we estimate shared loadings $\{\overline{\bU}_m\}_{m=1}^3$ using a HOSVD initialization followed by HOOI refinement operated on the mode-$m$ sample covariance. We intentionally overspecify the Tucker ranks and set $\overline{\bR} = (\overline{R}_1, \overline{R}_2, \overline{R}_3) = (4, 4, 4)$, producing estimated cores $\widehat{\mathcal{C}}_i = \widetilde{\mathcal{X}}_i \times_1 \overline{\bU}_1^{\top} \times_2 \overline{\bU}_2^{\top} \times_3 \overline{\bU}_3^{\top} \in \mathbb{R}^{4 \times 4 \times 4}$ for all splits. In the CP-core estimation, we estimate shared CP factors via a cPCA initialization followed by an ISO refinement operated on the mode-$m$ sample covariance. We overspecify the CP rank to $\overline{R} = 16$, and after factor estimation we compute per-sample CP coefficients by contracting $\widetilde{\mathcal{X}}_i$ with the pseudo-inverse loadings, embedding the resulting coefficient vector into a super-diagonal core $\widehat{\mathcal{C}}_i$.

Given the estimated cores $\widehat{\mathcal{C}}_i$, we train a DC-TNN classifier $\widehat{\pi}(\mathcal{X}_i) = \widehat{f}(\widehat{\mathcal{C}}_i, \widehat{\mathcal{U}}_{D,i})$ as in \eqref{eqn:TTN-penalty} with learned sparse selector $\cW_u$ that produces a low-dimensional refinement representation $\widehat{\mathcal{U}}_{D,i} = \hat \cW_u \bullet \tilde \cX_i$. We set the refinement-channel shape to $(\overline{K}_1, \overline{K}_2, \overline{K}_3) = (3, 3, 3)$ with $\overline{K}_{\text{tot}} = 27$, network depth $L = 3$, and we apply ReLU activations with layer normalization in both channels. We optimize the binary cross-entropy loss plus the clipped-$\ell_1$ selector penalty $\sum_{\cI} \rho_{\lambda}((\cW_u)_{\cI})$ with clipping threshold $\tau = 0.05$ and penalty level $\lambda = 0.1$, and we train with the Adam optimizer using learning rate $10^{-3}$, weight decay $10^{-4}$, batch size $128$.

\noindent\underline{\textbf{Compare with other models.}}
To benchmark the proposed core-refinement DC-TNN against tensor neural networks (TNNs) that exploit low-rank structure in alternative ways, we implement two additional TNNs that output $\widehat{\pi}(\tilde{\cX}_i) = \sigma(z_i)$. The first comparison model is imposing low-rank structure on weight tensors in Tensor Regression Layer (Low-rank TRL), each TRL is also followed by a ReLU activation, namely $\cH^{(0)} = \tilde{\cX}_i$, and $\cH^{(\ell+1)} = \text{ReLU}\{\text{TRL}(\cH^{(\ell)})\}$ for $\ell = 0, \ldots, L-1$, the weight tensor is factorized within the layer using either a Tucker or CP parameterization; to reduce computation, the hidden tensor widths are allowed to decay across depth (e.g., $32 \to 16 \to 8 \to 4$), after which we vectorize the final hidden tensor and apply a single linear projection to produce the logit $z_i$ and the probability $\widehat{\pi}(\tilde{\cX}_i)$. The second comparison model enforces low-rank structure across depth by stacking the hidden-layer regression weight tensors into a single ``whole'' tensor $\cW$ whose first mode indexes layers, so that each hidden layer uses the slice $\cW^{(\ell)} = \cW[\ell, :]$ in the same tensor regression update $\cH^{(\ell+1)} = \text{ReLU}\{\text{TRL}(\cH^{(\ell)})\}$ while keeping all hidden layers shape-preserving at $(32, 32, 32)$; we impose either a CP or Tucker factorization on $\cW$ (Depth-CP or Depth-Tucker). For the output layer (which is excluded from the stacking constraint), we use adaptive average pooling with $k = 6$, then apply a single linear projection to produce the logit $z_i$ and the probability $\widehat{\pi}(\tilde{\cX}_i)$. Following the same experimental design, we fit both the Low-rank TRL network and the stacked-weight network under Tucker and CP factorizations on both Tucker-generated and CP-generated datasets, yielding four model-data combinations for each baseline. We use identical train/calibration/test splits and the same BCE-based training to ensure fair comparison.

\begin{table}[h]
\centering
\caption{Estimation Results on Tucker-Generated Data}
\label{tab:tucker_results}
\resizebox{\textwidth}{!}{
\begin{tabular}{lcccccc}
\hline
Method & Test Acc & MSE (Test) & $\bar{\pi}|y{=}1$ (Cal) & $\bar{\pi}|y{=}1$ (Test) & $\bar{\pi}|y{=}0$ (Cal) & $\bar{\pi}|y{=}0$ (Test) \\
\hline
\cellcolor{gray!20}DC-TNN (Tucker) & \cellcolor{gray!20}0.853 & \cellcolor{gray!20}0.040 & $0.699_{(0.215)}$ & $0.734_{(0.167)}$ & $0.321_{(0.259)}$ & $0.292_{(0.231)}$ \\
DC-TNN (CP) & 0.825 & 0.066 & $0.680_{(0.225)}$ & $0.705_{(0.204)}$ & $0.341_{(0.235)}$ & $0.331_{(0.226)}$ \\
Stacked-TNN (Tucker) & 0.798 & 0.078 & $0.699_{(0.278)}$ & $0.746_{(0.231)}$ & $0.324_{(0.304)}$ & $0.312_{(0.290)}$ \\
Stacked-TNN (CP) & 0.780 & 0.083 & $0.692_{(0.266)}$ & $0.720_{(0.241)}$ & $0.317_{(0.292)}$ & $0.332_{(0.283)}$ \\
TRL-TNN (Tucker) & 0.788 & 0.093 & $0.708_{(0.243)}$ & $0.697_{(0.246)}$ & $0.351_{(0.275)}$ & $0.337_{(0.262)}$ \\
TRL-TNN (CP) & 0.805 & 0.082 & $0.697_{(0.263)}$ & $0.720_{(0.245)}$ & $0.326_{(0.274)}$ & $0.319_{(0.270)}$ \\
\hline
\end{tabular}
}
\end{table}

\begin{table}[h]
\centering
\caption{Estimation Results on CP-Generated Data}
\label{tab:cp_results}
\resizebox{\textwidth}{!}{
\begin{tabular}{lccccccc}
\hline
Method & Test Acc & MSE (Test) & $\bar{\pi}|y{=}1$ (Cal) & $\bar{\pi}|y{=}1$ (Test) & $\bar{\pi}|y{=}0$ (Cal) & $\bar{\pi}|y{=}0$ (Test) \\
\hline
\cellcolor{gray!20}DC-TNN (CP) & \cellcolor{gray!20}0.855 & \cellcolor{gray!20}0.024 & $0.800_{(0.266)}$ & $0.784_{(0.266)}$ & $0.223_{(0.227)}$ & $0.222_{(0.212)}$ \\
DC-TNN (Tucker) & 0.805 & 0.058 & $0.671_{(0.195)}$ & $0.666_{(0.196)}$ & $0.306_{(0.161)}$ & $0.330_{(0.184)}$ \\
Stacked-TNN (CP) & 0.728 & 0.117 & $0.671_{(0.310)}$ & $0.670_{(0.326)}$ & $0.338_{(0.316)}$ & $0.324_{(0.292)}$ \\
Stacked-TNN (Tucker) & 0.715 & 0.133 & $0.733_{(0.351)}$ & $0.633_{(0.389)}$ & $0.243_{(0.328)}$ & $0.250_{(0.330)}$ \\
TRL-TNN (CP) & 0.730 & 0.110 & $0.689_{(0.302)}$ & $0.671_{(0.319)}$ & $0.334_{(0.264)}$ & $0.338_{(0.266)}$ \\
TRL-TNN (Tucker) & 0.780 & 0.081 & $0.733_{(0.281)}$ & $0.706_{(0.296)}$ & $0.278_{(0.279)}$ & $0.285_{(0.273)}$ \\
\hline
\end{tabular}
}
\end{table}

\begin{figure}[h]
\centering
\includegraphics[width=0.9\textwidth]{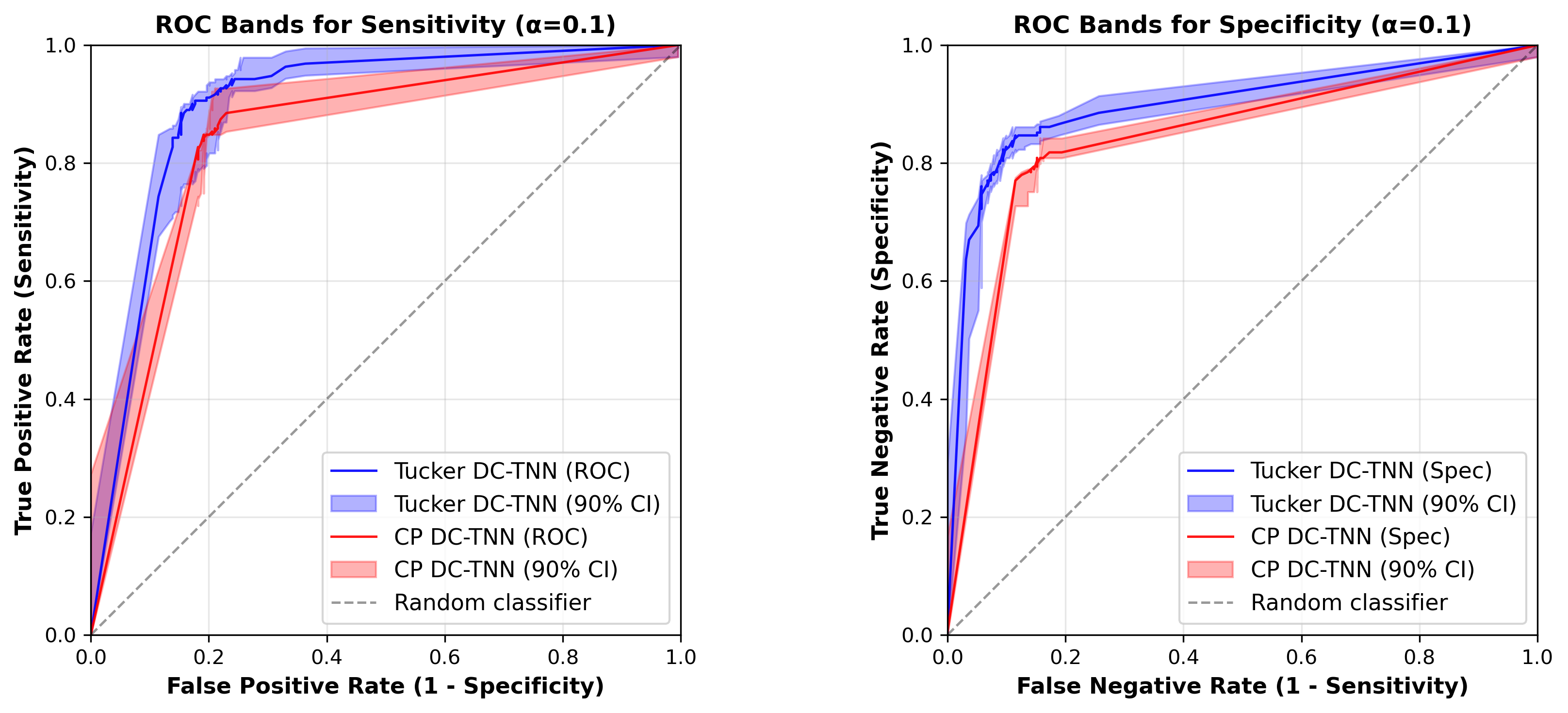}
\caption{\small ROC bands with 90\% conformal confidence intervals for Tucker-generated data.}
\label{fig:roc_bands_tucker}
\end{figure}

\begin{figure}[h]
\centering
\includegraphics[width=0.9\textwidth]{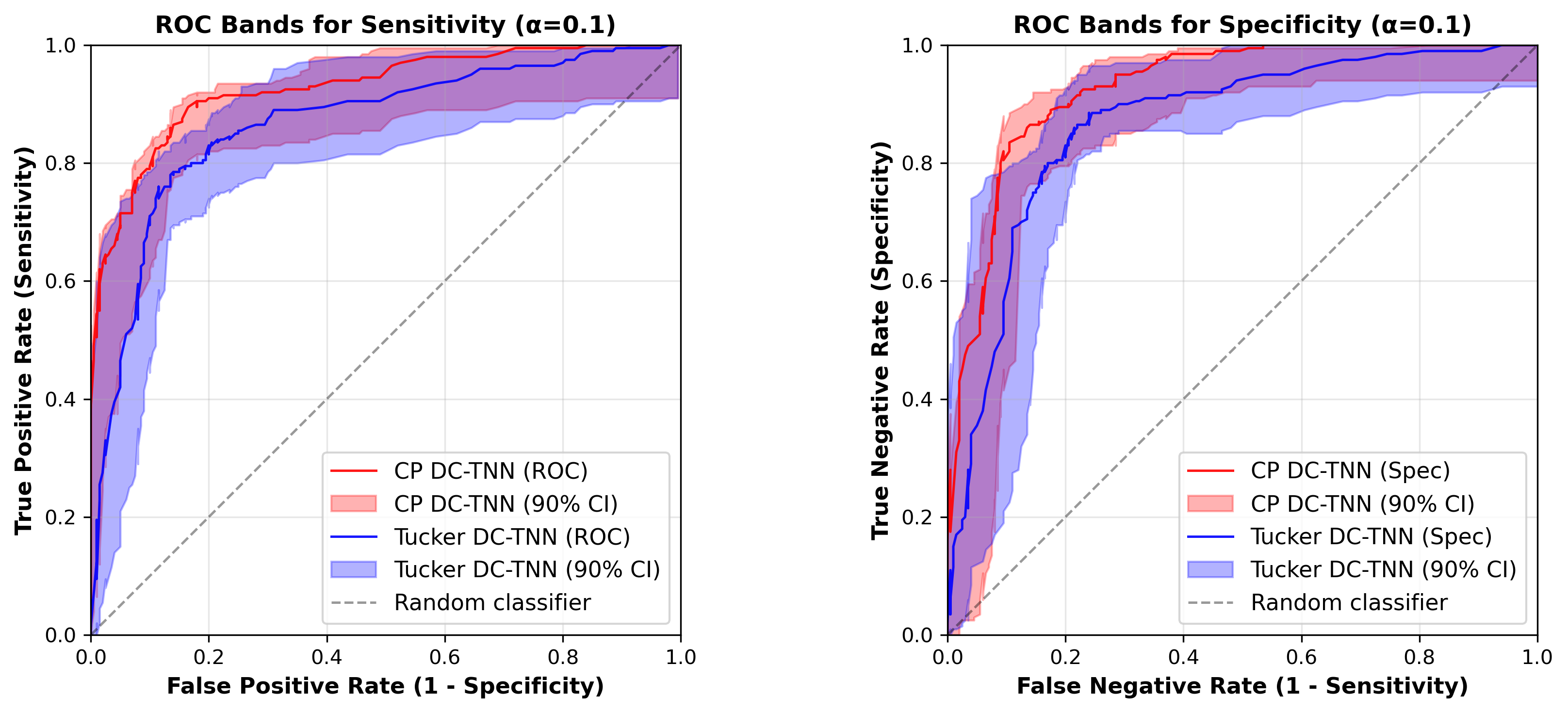}
\caption{\small ROC bands with 90\% conformal confidence intervals for CP-generated data.}
\label{fig:roc_bands_cp}
\end{figure}

\noindent\underline{\textbf{Uncertainty Quantification.}} We quantify predictive uncertainty using the conformal prediction ROC band procedure in Section~\ref{sec:uq}. For each fitted model, we employ $K_{\text{tr}} = 50$ neighbors for KNN regression to approximate $\pi(\cdot)$ using predicted probabilities from the training split and $K_{\text{ca}} = 10$ nearest calibration points for local calibration sets. With miscoverage level $\alpha = 0.1$, we construct class-conditional ROC bands and AUC confidence intervals over $G = 200$ thresholds. Figures~\ref{fig:roc_bands_tucker} and \ref{fig:roc_bands_cp} display the ROC bands for DC-TNN methods on Tucker-generated and CP-generated data respectively. Tables~\ref{tab:uq_auc_tucker} and \ref{tab:uq_auc_cp} report AUC point estimates and 90\% confidence intervals for all six methods on both data generation regimes, validating the coverage guarantees of Proposition~\ref{prop: cov CP_ROC} and Corollary~\ref{cor: AUC confidence band}.

\begin{table}[h]
\centering
\caption{AUC Point Estimates and 90\% Conformal Confidence Intervals on Tucker-Generated Data}
\label{tab:uq_auc_tucker}
\resizebox{0.8\textwidth}{!}{
\begin{tabular}{lcccc}
\hline
Method & AUC (Sens) & CI (Sens) & AUC (Spec) & CI (Spec) \\
\hline
\cellcolor{gray!20}DC-TNN (Tucker) & \cellcolor{gray!20}0.838 & \cellcolor{gray!20}[0.768, 0.882] & \cellcolor{gray!20}0.843 & \cellcolor{gray!20}[0.805, 0.878] \\
DC-TNN (CP) & 0.718 & [0.693, 0.801] & 0.723 & [0.713, 0.762] \\
Stacked-TNN (Tucker) & 0.829 & [0.727, 0.905] & 0.834 & [0.753, 0.917] \\
Stacked-TNN (CP) & 0.806 & [0.684, 0.927] & 0.805 & [0.736, 0.911] \\
TRL-TNN (Tucker) & 0.795 & [0.682, 0.894] & 0.800 & [0.695, 0.899] \\
TRL-TNN (CP) & 0.786 & [0.689, 0.922] & 0.791 & [0.702, 0.897] \\
\hline
\end{tabular}
}
\end{table}

\begin{table}[h]
\centering
\caption{\small AUC Point Estimates and 90\% Conformal Confidence Intervals on CP-Generated Data}
\label{tab:uq_auc_cp}
\resizebox{0.8\textwidth}{!}{
\begin{tabular}{lcccc}
\hline
Method & AUC (Sens) & CI (Sens) & AUC (Spec) & CI (Spec) \\
\hline
\cellcolor{gray!20}DC-TNN (CP) & \cellcolor{gray!20}0.921 & \cellcolor{gray!20}[0.788, 0.941] & \cellcolor{gray!20}0.925 & \cellcolor{gray!20}[0.791, 0.944] \\
DC-TNN (Tucker) & 0.858 & [0.732, 0.927] & 0.864 & [0.751, 0.941] \\
Stacked-TNN (CP) & 0.759 & [0.559, 0.931] & 0.764 & [0.553, 0.931] \\
Stacked-TNN (Tucker) & 0.758 & [0.523, 0.906] & 0.763 & [0.522, 0.920] \\
TRL-TNN (CP) & 0.764 & [0.539, 0.910] & 0.768 & [0.571, 0.902] \\
TRL-TNN (Tucker) & 0.843 & [0.645, 0.919] & 0.848 & [0.623, 0.931] \\
\hline
\end{tabular}
}
\end{table}

\begin{figure}[h]
\centering
\includegraphics[width=0.8\textwidth]{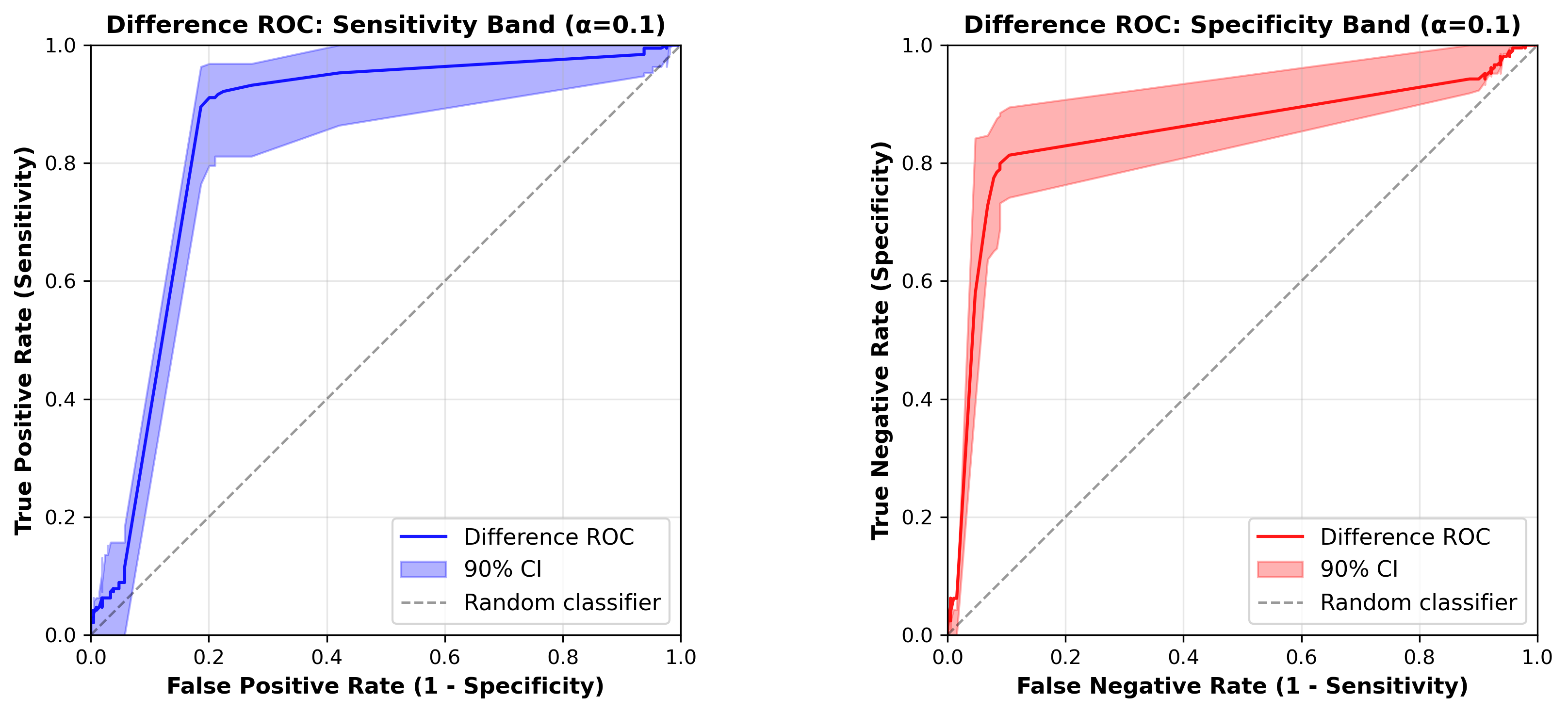}
\caption{\small Difference ROC bands with 90\% conformal confidence intervals for Tucker-generated data.}
\label{fig:diff_roc_tucker}
\end{figure}

\begin{figure}[H]
\centering
\includegraphics[width=0.8\textwidth]{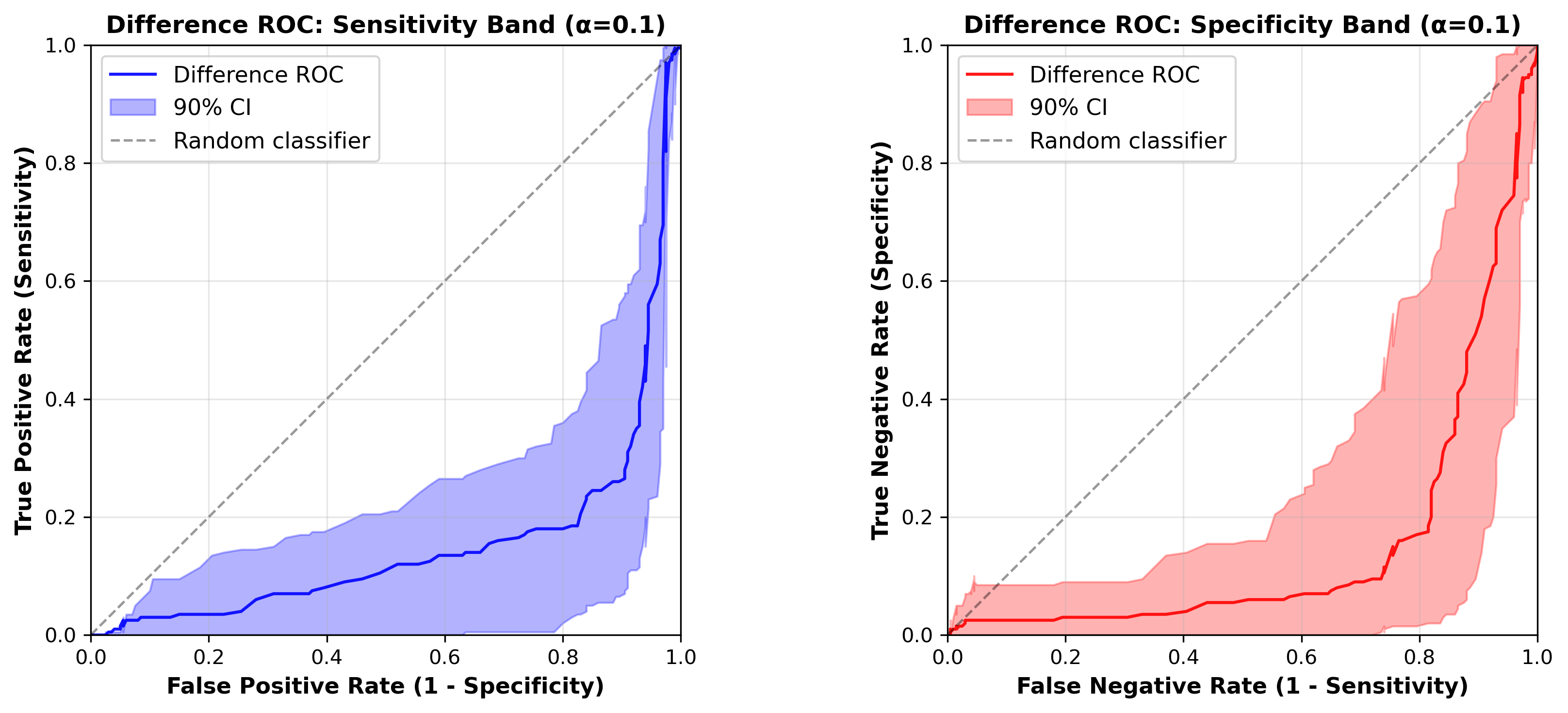}
\caption{\small Difference ROC bands with 90\% conformal confidence intervals for CP-generated data.}
\label{fig:diff_roc_cp}
\end{figure}

\noindent\underline{\textbf{Structure Selection.}} We evaluate the conformal ROC-based structure selector in Section~\ref{sec:selector} by comparing Tucker-core and CP-core DC-TNNs on both data generation regimes. We define the difference score $d(\mathcal{X}) = \widehat{\pi}^{\text{Tucker}}(\mathcal{X}) - \widehat{\pi}^{\text{CP}}(\mathcal{X}) \in [-1, 1]$ and construct class-conditional local calibration neighborhoods of size $K = 8$ using the weighted core-refinement distance with $w = 10$. Since each model produces its own latent representation, we compute distances in the Tucker latent space, which aligns with the null hypothesis that Tucker provides the better structure. Following Algorithm~\ref{algo:Selector}, we form prediction intervals at level $\alpha = 0.1$ and aggregate them into difference-ROC bands over $G = 200$ threshold values. We compute sensitivity-based and specificity-based AUC confidence intervals and apply the decision rule from Section~\ref{subsec:selector}. Figures~\ref{fig:diff_roc_tucker} and \ref{fig:diff_roc_cp} display the difference ROC bands, and Table~\ref{tab:selector_decisions} reports the selection decisions. The results validate the coverage guarantees of Theorem~\ref{thm:diff-roc-fs} and demonstrate the consistency property of Theorem~\ref{thm: selector consistency}, with both methods correctly identifying the true underlying structure in their respective regimes.

\begin{table}[h]
\centering
\caption{\small Structure Selection Results via Difference ROC}
\label{tab:selector_decisions}
\resizebox{0.95\textwidth}{!}{
\begin{tabular}{lccc}
\hline
Data & AUC$_{\text{Sens}}$ [90\% CI] & AUC$_{\text{Spec}}$ [90\% CI] & Decision \\
\hline
Tucker-generated dataset & 0.785 [0.692, 0.831] & 0.790 [0.723, 0.864] & DC-TNN (Tucker) \\
CP-generated dataset & 0.149 [0.040, 0.269] & 0.149 [0.047, 0.329] & DC-TNN (CP) \\
\hline
\end{tabular}
}
\end{table}

\section{Real Data Analysis}\label{sec:real-data}
We apply the proposed framework to the \texttt{DD} benchmark from \texttt{TUDataset}, a collection of graphs representing protein structures, where each observation is a graph $G$ with a binary label $y \in \{0, 1\}$ indicating enzyme or non-enzyme status. 
Our analysis pursues two objectives: (i) benchmarking the core-refinement DC-TNN against alternative low-rank tensor neural networks, and (ii) selecting between Tucker-type and CP-type low-rank structure in a data-driven, statistically valid way.

Since the proposed methods operate on tensor covariates, we first map each graph to a fixed-dimensional tensor feature. Let $\mathrm{PI}(G)$ denote the persistence image representation of $G$ (construction details are deferred to the Appendix). An encoder
$(G, \mathrm{PI}(G)) \mapsto \mathcal{X} \in \mathbb{R}^{D_1 \times D_2 \times D_3}$ produces the tensor feature via two branches: a graph branch aggregating node and edge 
information, and a PI branch extracting multiscale topological signals, followed by a fusion module. The same encoder architecture and hyperparameters are used for all competing 
methods to ensure fair comparison; full details appear in the Appendix.

\noindent\underline{\textbf{Implementation.}}
We follow the same stratified train/calibration/test split strategy as in Section~\ref{sec:simulation}, denoting the resulting index sets by $\mathcal{I}_{\mathrm{tr}}$, $\mathcal{I}_{\mathrm{ca}}$, and $\mathcal{I}_{\mathrm{te}}$. 
For each method, the encoder and the downstream tensor network classifier are trained \emph{jointly} end-to-end. To ensure fair comparison, all competing methods share the same encoder \emph{architecture} but are each initialized from scratch with their own encoder instance. 

We compare two DC-TNN variants against the TRL-TNN and Stacked-weight TNN baselines introduced in Section~\ref{sec:simulation}. In the Tucker-core and CP-core DC-TNN, the encoder output $\mathcal{X}$ is further decomposed into a low-rank core $\widehat{\mathcal{C}}$ via the respective factorization, and the dual-channel architecture processes 
$(\widehat{\mathcal{C}}, \mathcal{X})$ jointly. The TRL-TNN and Stacked-weight TNN baselines map $\mathcal{X}$ directly to the 
predicted probability through their respective single-channel architectures. All models are trained under the same optimization scheme and stopping rule, and we report test accuracy 
together with summary statistics of predicted probabilities on both the calibration and test splits.

\begin{table}[t]
\centering
\caption{Estimation results on the \texttt{DD} dataset. Values reported as mean (std).}
\label{tab:dd-estimation}
\resizebox{\textwidth}{!}{\begin{tabular}{lcccccc}
\hline
Method & Test Acc & $\overline{\pi}\,|\,y{=}1$ (Cal) & $\overline{\pi}\,|\,y{=}1$ (Test) & $\overline{\pi}\,|\,y{=}0$ (Cal) & $\overline{\pi}\,|\,y{=}0$ (Test) \\
\hline
DC-TNN (Tucker)      & 0.803 & 0.638 (0.281) & 0.618 (0.256) & 0.265 (0.160) & 0.270 (0.174) \\
DC-TNN (CP)          & 0.829 & 0.674 (0.263) & 0.670 (0.272) & 0.208 (0.202) & 0.243 (0.196) \\
Stacked-TNN (Tucker) & 0.761 & 0.705 (0.269) & 0.752 (0.224) & 0.333 (0.287) & 0.333 (0.270) \\
Stacked-TNN (CP)     & 0.761 & 0.504 (0.325) & 0.523 (0.334) & 0.156 (0.171) & 0.162 (0.166) \\
TRL-TNN (Tucker)     & 0.786 & 0.654 (0.309) & 0.661 (0.299) & 0.236 (0.264) & 0.275 (0.267) \\
TRL-TNN (CP)         & 0.769 & 0.583 (0.310) & 0.601 (0.193) & 0.239 (0.195) & 0.340 (0.210) \\
\hline
\end{tabular}
}
\end{table}

\begin{figure}[t]
\centering
\resizebox{0.85\textwidth}{!}{\begin{subfigure}[b]{0.48\linewidth}
    \centering
    \includegraphics[width=\linewidth]{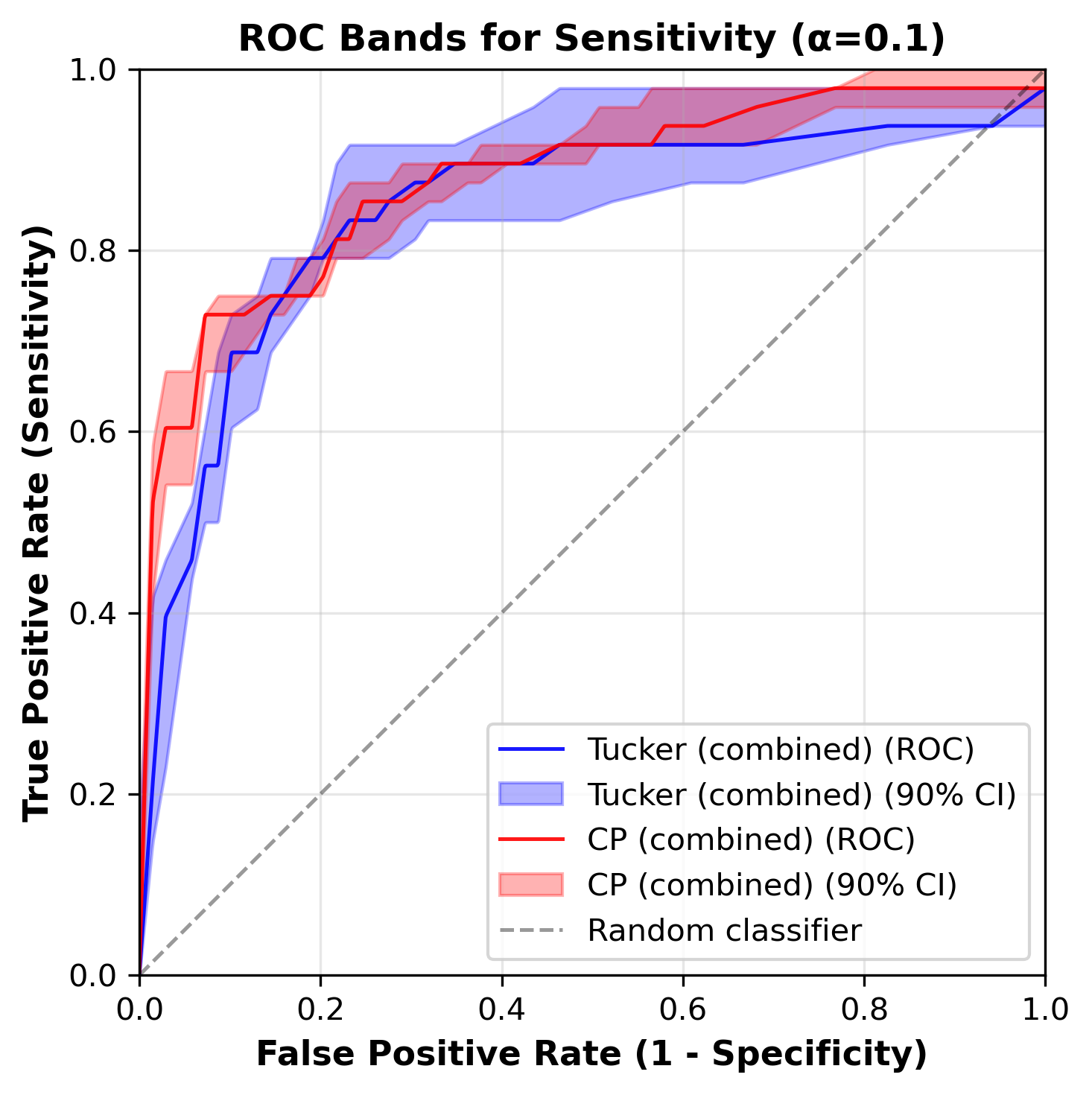}
    \caption{Sensitivity ROC bands.}
    \label{fig:dd-roc-sens}
\end{subfigure}
\hfill
\begin{subfigure}[b]{0.48\linewidth}
    \centering
    \includegraphics[width=\linewidth]{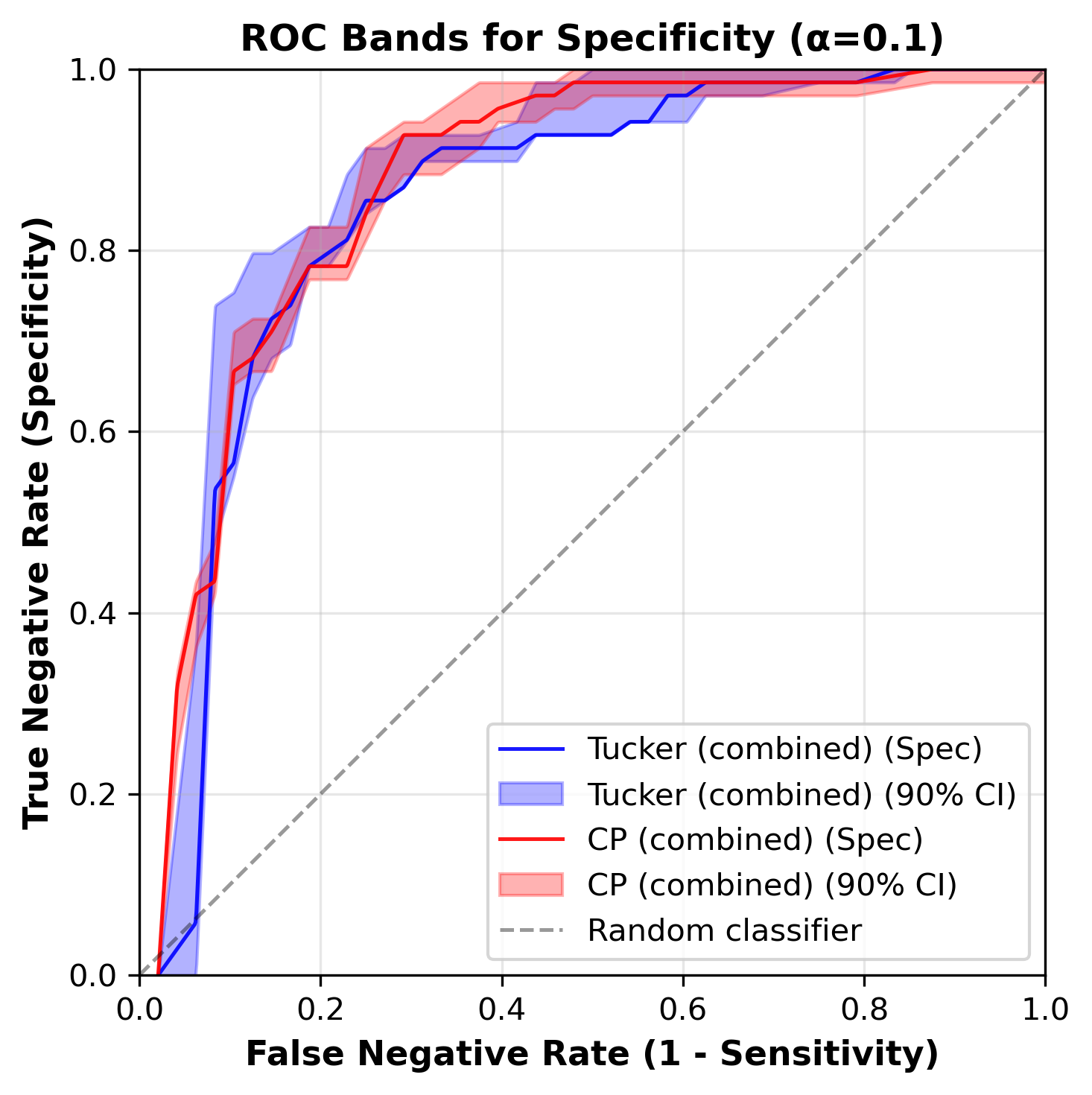}
    \caption{Specificity ROC bands.}
    \label{fig:dd-roc-spec}
\end{subfigure}
}
\caption{ROC bands with 90\% conformal confidence intervals for DC-TNN methods 
on the \texttt{DD} dataset ($\alpha = 0.1$).}
\label{fig:dd-roc-bands}
\end{figure}

\noindent\underline{\textbf{Uncertainty Quantification.}}
We quantify predictive uncertainty using the conformal ROC band procedure from 
Section~\ref{sec:uq}. For each fitted model, we construct class-conditional ROC bands 
at miscoverage level $\alpha = 0.1$ via KNN regression to approximate $\pi(\cdot)$ on 
the training split and local calibration sets formed from the calibration split. 
Figure~\ref{fig:dd-roc-bands} displays the resulting sensitivity and specificity ROC 
bands for the DC-TNN methods, and Table~\ref{tab:dd-auc} reports AUC point estimates 
with conformal confidence intervals for all methods.

\begin{table}[t]
\centering
\caption{AUC point estimates and 90\% conformal confidence intervals on the 
\texttt{DD} dataset.}
\label{tab:dd-auc}
\vspace{2mm}
\resizebox{0.9\textwidth}{!}{\begin{tabular}{lcccc}
\hline
Method & AUC (Sens) & CI (Sens) & AUC (Spec) & CI (Spec) \\
\hline
DC-TNN (Tucker)      & 0.838 & [0.785, 0.893] & 0.836 & [0.823, 0.885] \\
DC-TNN (CP)          & 0.870 & [0.845, 0.899] & 0.868 & [0.847, 0.893] \\
Stacked-TNN (Tucker) & 0.817 & [0.743, 0.842] & 0.822 & [0.795, 0.884] \\
Stacked-TNN (CP)     & 0.805 & [0.541, 0.833] & 0.818 & [0.792, 0.884] \\
TRL-TNN (Tucker)     & 0.813 & [0.731, 0.863] & 0.819 & [0.787, 0.867] \\
TRL-TNN (CP)         & 0.781 & [0.714, 0.803] & 0.790 & [0.682, 0.868] \\
\hline
\end{tabular}
}
\end{table}

\noindent\underline{\textbf{Structure Selection.}}
We apply the conformal ROC-based structure selector from Section~\ref{sec:selector} to determine which low-rank form is more strongly supported by the \texttt{DD} data. Since the DC-TNN explicitly decomposes each tensor covariate $\mathcal{X}$ into a low-rank core $\widehat{\mathcal{C}}$ and a sparse refinement $\widehat{\mathcal{U}}$, the structure selection is carried out exclusively for the DC-TNN variants and is framed as a two-directional conformal hypothesis test. Structure selection results for the competing model, TRL-TNN and Stacked-TNN, are reported in Appendix~\ref{sec:additional results}.

\begin{figure}[t]
\centering
\resizebox{0.85\textwidth}{!}{
\begin{subfigure}[b]{0.48\linewidth}
    \centering
    \includegraphics[width=\linewidth]{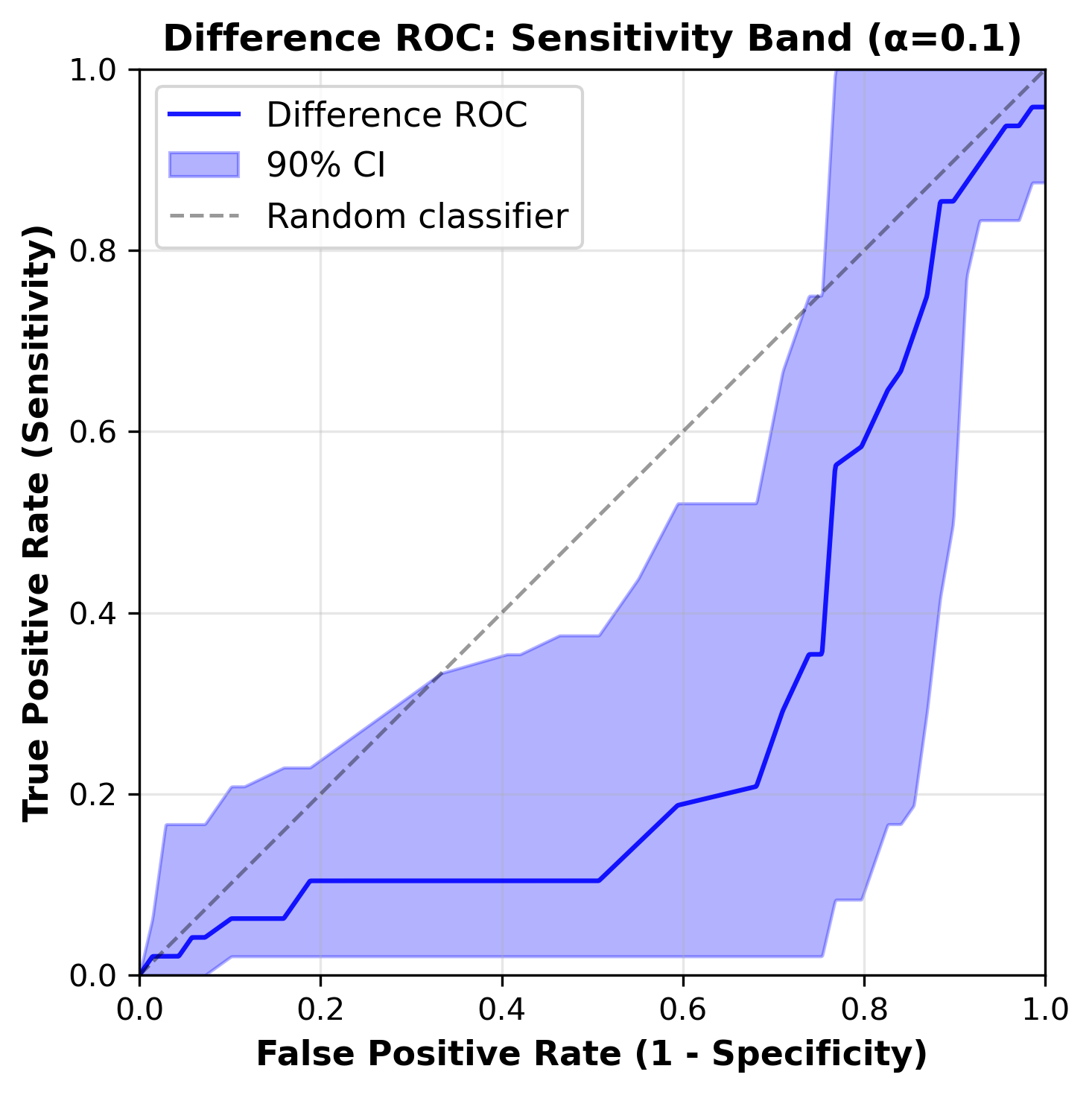}
    \caption{Sensitivity difference ROC band.}
    \label{fig:dd-diff-roc-sens-1}
\end{subfigure}
\hfill
\begin{subfigure}[b]{0.48\linewidth}
    \centering
    \includegraphics[width=\linewidth]{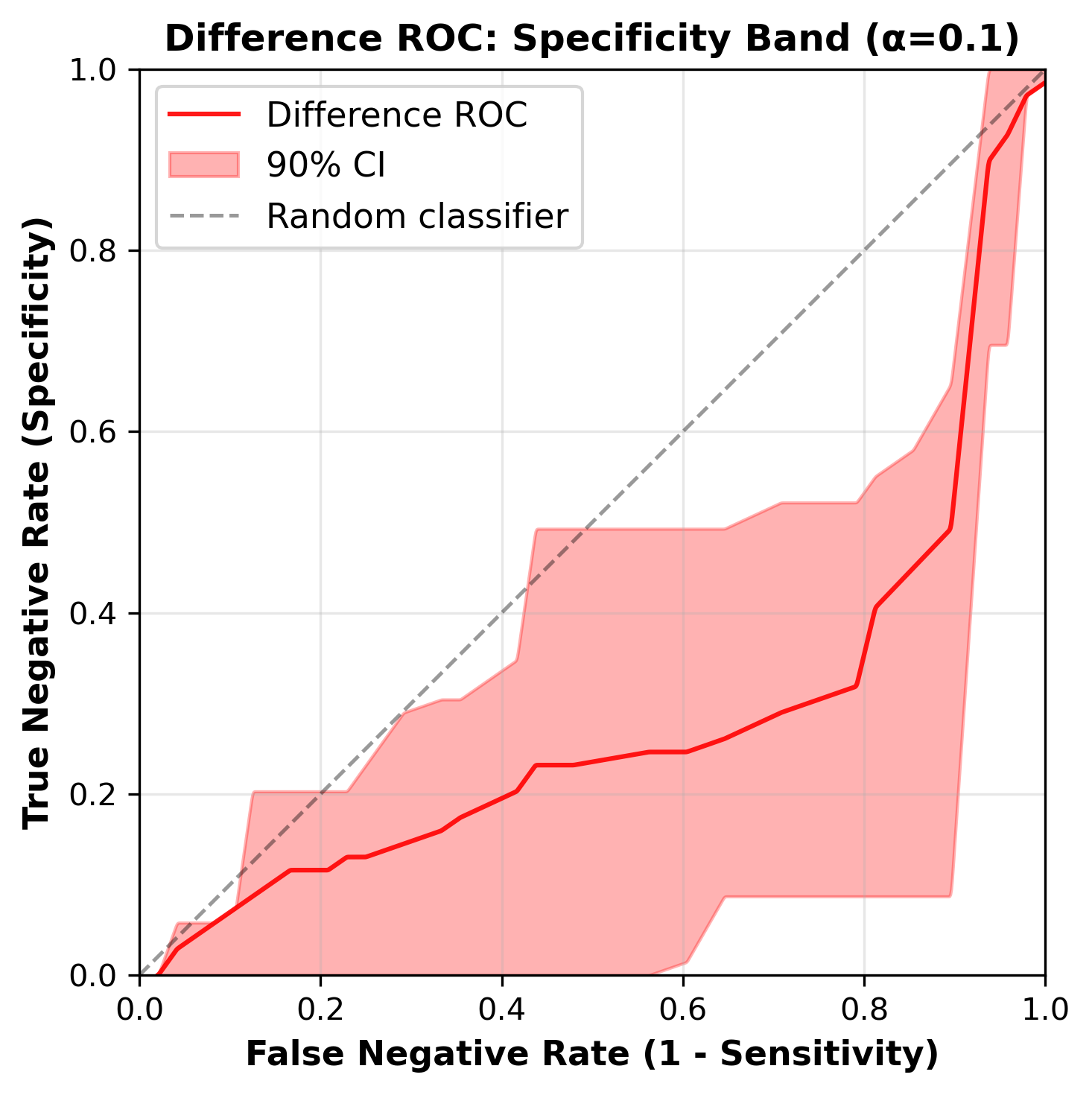}
    \caption{Specificity difference ROC band.}
    \label{fig:dd-diff-roc-spec-1}
\end{subfigure}
}
\caption{Test 1 ($H_0^{(1)}$): Difference ROC bands with 90\% conformal confidence 
intervals for $d^{(1)}(\mathcal{X}) = \widehat{\pi}_{\mathrm{Tucker}}(\mathcal{X}) - 
\widehat{\pi}_{\mathrm{CP}}(\mathcal{X})$ on the \texttt{DD} dataset ($\alpha = 0.1$).}
\label{fig:dd-diff-roc-1}
\end{figure}

\begin{figure}[H]
\centering
\resizebox{0.85\textwidth}{!}{
\begin{subfigure}[b]{0.48\linewidth}
    \centering
    \includegraphics[width=\linewidth]{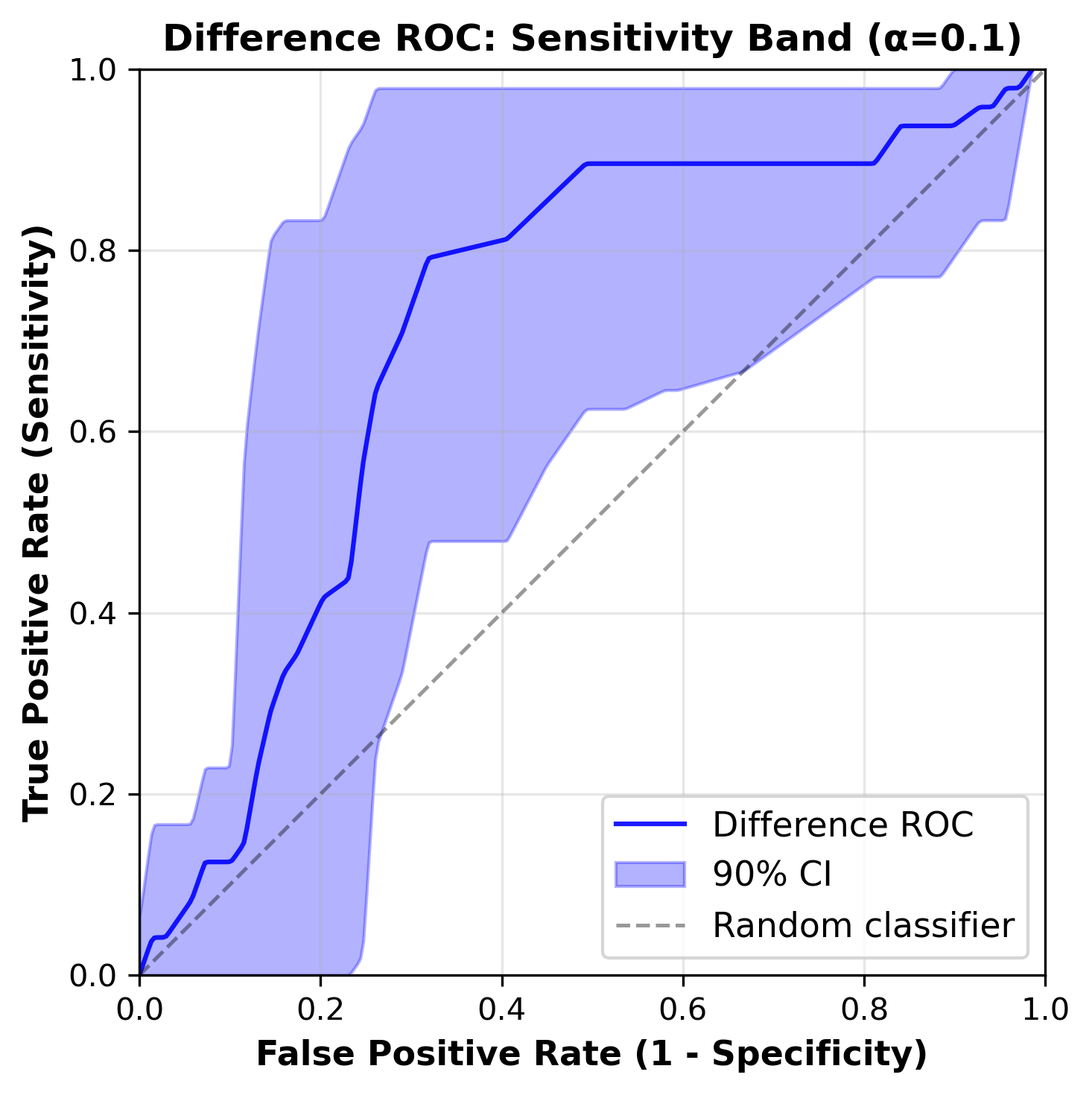}
    \caption{Sensitivity difference ROC band.}
    \label{fig:dd-diff-roc-sens-2}
\end{subfigure}
\hfill
\begin{subfigure}[b]{0.48\linewidth}
    \centering
    \includegraphics[width=\linewidth]{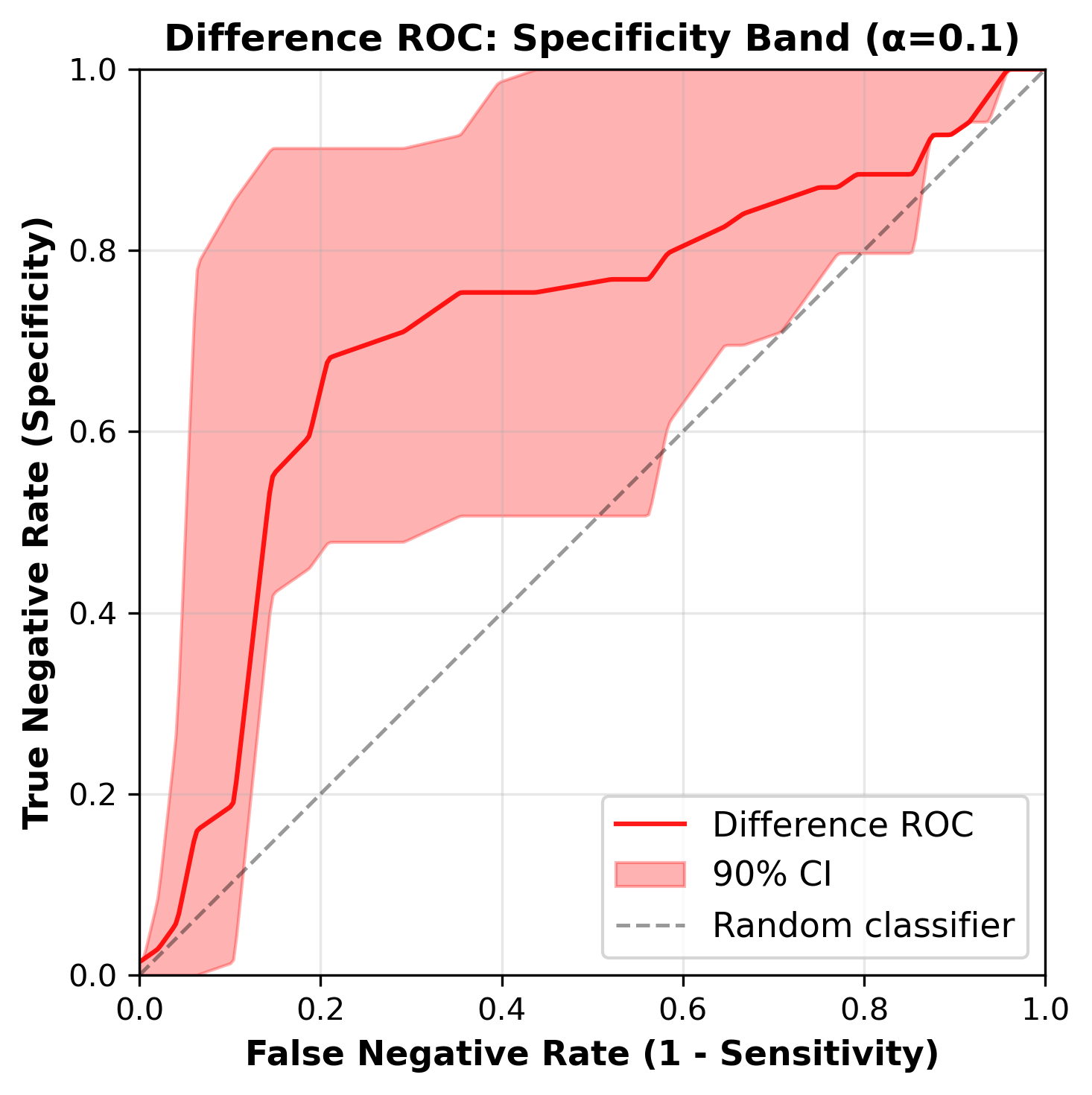}
    \caption{Specificity difference ROC band.}
    \label{fig:dd-diff-roc-spec-2}
\end{subfigure}
}
\caption{Test 2 ($H_0^{(2)}$): Difference ROC bands with 90\% conformal confidence 
intervals for $d^{(2)}(\mathcal{X}) = \widehat{\pi}_{\mathrm{CP}}(\mathcal{X}) - 
\widehat{\pi}_{\mathrm{Tucker}}(\mathcal{X})$ on the \texttt{DD} dataset ($\alpha = 0.1$).}
\label{fig:dd-diff-roc-2}
\end{figure}

We first test whether the Tucker factor yields better 
predictions than CP. The null hypothesis is $H_0^{(1)}: \hat \pi^{(Tucker)}\text{ outperforms } \hat \pi^{(CP)}$ against $H_a^{(1)}: \hat \pi^{(Tucker)} \text{ no better than } \hat \pi^{(CP)}$. Specifically, we compute the difference score $d^{(1)}(\mathcal{X}) = \hat\pi_{\mathrm{Tucker}}(\mathcal{X}) - \hat\pi_{\mathrm{CP}}(\mathcal{X}) \in [-1,1]$, construct local neighborhoods using the core-refinement distance $d_{\mathrm{FA}}$ in~\eqref{eqn:fa-distance}, which measures similarity jointly in the latent core space 
$\widehat{\mathcal{C}}^{Tucker}$ and the sparse refinement space $\widehat{\mathcal{U}}^{Tucker}$, and form conformal difference ROC bands at miscoverage level $\alpha = 0.1$. To guard against directional bias, we additionally test the complementary null $H_0^{(2)}: \hat \pi^{(CP)}\text{ outperforms } \hat \pi^{(Tucker)}$ by reversing the difference score to 
$d^{(2)}(\mathcal{X}) = \widehat{\pi}_{\mathrm{CP}}(\mathcal{X}) - \widehat{\pi}_{\mathrm{Tucker}}(\mathcal{X})$, 
and changing to the CP latent space for local neighborhoods. Table~\ref{tab:dd-structure} reports the results. Since $H_0^{(1)}$ is rejected (both AUC CIs lie entirely below 0.5) while $H_0^{(2)}$ yields a tie (CIs straddle 0.5), we conclude that Tucker does not outperform CP but CP may outperform Tucker. Those two consistent tests leads to the final selection of DC-TNN (CP). Figure~\ref{fig:dd-diff-roc-1} and Figure~\ref{fig:dd-diff-roc-2} display the corresponding difference ROC bands for $H_0^{(1)}$ and $H_0^{(2)}$, respectively.

\begin{table}[H]
\centering
\caption{Structure selection results on the \texttt{DD} dataset via conformal difference ROC.}
\label{tab:dd-structure}
\vspace{2mm}
\resizebox{\textwidth}{!}{\begin{tabular}{lccc}
\hline
Test (Null $H_0$) & AUC$_{\mathrm{sens}}$ [90\% CI] & AUC$_{\mathrm{spec}}$ [90\% CI] & Decision \\
\hline
$H_0^{(1)}: \hat \pi^{(Tucker)}\text{ outperforms } \hat \pi^{(CP)}$ & 0.270 [0.121, 0.495] & 0.271 [0.068, 0.421] & Reject $\Rightarrow$ CP \\
$H_0^{(2)}: \hat \pi^{(CP)}\text{ outperforms } \hat \pi^{(Tucker)}$ & 0.694 [0.453, 0.856] & 0.704 [0.558, 0.895] & Tie \\
\hline
\multicolumn{4}{l}{\textbf{Final decision: DC-TNN (CP)}} \\
\hline
\end{tabular}
}
\end{table}

\section{Conclusion}
\label{sec:conc}

We developed a unified framework for tensor regression in which a single core–refinement representation supports estimation, distribution-free inference, and structure selection. The Dual-Channel Tensor Neural Network decomposes each tensor input into a low-rank latent core and a sparse refinement, processes them through coupled neural channels, and accommodates Tucker, CP, and tensor-train cores within a common architecture. We established non-asymptotic risk bounds whose effective dimension depends on the core rank and refinement sparsity rather than the ambient tensor size, developed structure-aware conformal ROC and AUC bands that calibrate locally in the latent space with finite-sample coverage, and introduced what is, to our knowledge, the first distribution-free procedure with finite-sample validity for selecting among candidate low-rank tensor decompositions.

Several extensions are natural. The conformal selector generalizes to more than two candidate structures via simultaneous difference-ROC bands; a global theory for end-to-end joint estimation would complement the two-stage analysis developed here; and structure-aware calibration should extend to settings with covariate shift or temporal dependence, where ambient-space conformal methods are particularly conservative. More broadly, the results suggest that core-augmented inference may serve as a general organizing principle for statistical learning on tensor-valued data.

\spacingset{1.18}
\bibliographystyle{agsm}
\bibliography{bibmain}

@article{chen2023statistical,
  title={Statistical inference for high-dimensional matrix-variate factor models},
  author={Chen, Elynn Y. and Fan, Jianqing},
  journal={Journal of the American Statistical Association},
  volume={118},
  number={542},
  pages={1038--1055},
  year={2023},
  publisher={Taylor \& Francis},
  doi={10.1080/01621459.2021.1970569}
}

@article{liu1904helping,
  title={Helping effects against curse of dimensionality in threshold factor models for matrix time series},
  author={Liu, Xialu and Chen, Elynn Y.},
  journal={arXiv preprint arXiv:1904.07383},
  year={2019}
}

@article{chen2020constrained,
  title={Constrained factor models for high-dimensional matrix-variate time series},
  author={Chen, Elynn Y. and Tsay, Ruey S. and Chen, Rong},
  journal={Journal of the American Statistical Association},
  volume={115},
  number={530},
  pages={775--793},
  year={2020},
  publisher={Taylor \& Francis},
  doi={10.1080/01621459.2019.1584899}
}

@article{chen2022modeling,
  title={Modeling dynamic transport network with matrix factor models: with an application to international trade flow},
  author={Chen, Elynn Y. and Chen, Rong},
  journal={Journal of Data Science},
  volume={21},
  number={3},
  pages={490--507},
  year={2022},
  doi={10.6339/22-JDS1065}
}

@article{chen2020modeling,
  title={Modeling multivariate spatial-temporal data with latent low-dimensional dynamics},
  author={Chen, Elynn Y. and Yun, Xin and Chen, Rong and Yao, Qiwei},
  journal={arXiv preprint arXiv:2002.01305},
  year={2020}
}

@inproceedings{wen2024tensor,
  title={Tensor-view topological graph neural network},
  author={Wen, Tao and Chen, Elynn and Chen, Yuzhou},
  booktitle={Proceedings of the 27th International Conference on Artificial Intelligence and Statistics (AISTATS)},
  series={Proceedings of Machine Learning Research},
  volume={238},
  pages={4330--4338},
  year={2024},
  publisher={PMLR},
  address={Valencia, Spain}
}

@article{chen2024semi,
  title={Semi-parametric tensor factor analysis by iteratively projected singular value decomposition},
  author={Chen, Elynn Y. and Xia, Dong and Cai, Chencheng and Fan, Jianqing},
  journal={Journal of the Royal Statistical Society Series B: Statistical Methodology},
  volume={86},
  number={3},
  pages={793--823},
  year={2024},
  publisher={Oxford University Press},
  doi={10.1093/jrsssb/qkae001}
}

@article{chen2024time,
  title={Time-varying matrix factor models},
  author={Chen, Bin and Chen, Elynn Y. and Bolivar, Stevenson and Chen, Rong},
  journal={arXiv preprint arXiv:2404.01546},
  year={2024}
}

@article{chen2026factor,
  title={Factor augmented matrix regression},
  author={Chen, Elynn Y. and Fan, Jianqing and Zhu, Xiaonan},
  journal={Journal of the American Statistical Association},
  pages={1--14},
  year={2026},
  publisher={Taylor \& Francis},
  note={Published online December 2025}
}

@article{chen2024hightensorclass,
  title={High-dimensional tensor classification with {CP} low-rank discriminant structure},
  author={Chen, Elynn and Han, Yuefeng and Li, Jiayu},
  journal={arXiv preprint arXiv:2409.14397},
  year={2024}
}

@article{chen2024hightensordisc,
  title={High-dimensional tensor discriminant analysis with incomplete tensors},
  author={Chen, Elynn and Han, Yuefeng and Li, Jiayu},
  journal={arXiv preprint arXiv:2410.14783},
  year={2024}
}

@article{wu2024conditionalUQ,
  title={Conditional uncertainty quantification for tensorized topological neural networks},
  author={Wu, Yujia and Yang, Bo and Zhao, Yang and Chen, Elynn and Chen, Yuzhou and Zheng, Zheshi},
  journal={arXiv preprint arXiv:2410.15241},
  year={2024}
}

@inproceedings{kong2025teaformers,
  title={{TEAFormers}: Tensor-augmented transformers for multi-dimensional time series forecasting},
  author={Kong, Linghang and Chen, Elynn and Chen, Yuzhou and Han, Yuefeng},
  booktitle={IJCAI 2025 AI for Time Series Workshop},
  year={2025},
  note={arXiv:2410.20439}
}

@article{xu2025statistical,
  title={Statistical inference for low-rank tensor models},
  author={Xu, Ke and Chen, Elynn and Han, Yuefeng},
  journal={arXiv preprint arXiv:2501.16223},
  year={2025}
}

@inproceedings{wen2025bridging,
  title={Bridging domain adaptation and graph neural networks: A tensor-based framework for effective label propagation},
  author={Wen, Tao and Chen, Elynn and Chen, Yuzhou and Lei, Qi},
  booktitle={Conference on Parsimony and Learning (CPAL)},
  year={2025}
}

@inproceedings{wu2025tensor,
  title={Tensor-fused multi-view graph contrastive learning},
  author={Wu, Yujia and Mo, Junyi and Chen, Elynn and Chen, Yuzhou},
  booktitle={Advances in Knowledge Discovery and Data Mining (PAKDD)},
  series={Lecture Notes in Computer Science},
  volume={15876},
  pages={16--28},
  year={2025},
  publisher={Springer Nature Singapore},
  doi={10.1007/978-981-96-8298-0_2}
}

@inproceedings{wu2025conditional,
  title={Conditional prediction {ROC} bands for graph classification},
  author={Wu, Yujia and Yang, Bo and Chen, Elynn and Chen, Yuzhou and Zheng, Zheshi},
  booktitle={Proceedings of the 28th International Conference on Artificial Intelligence and Statistics (AISTATS)},
  series={Proceedings of Machine Learning Research},
  volume={258},
  pages={2458--2466},
  year={2025},
  publisher={PMLR}
}

@article{liu2025tensor,
  title={Tensor {Neyman-Pearson} classification: Theory, algorithms, and error control},
  author={Liu, Lingchong and Chen, Elynn and Han, Yuefeng and Xia, Lucy},
  journal={arXiv preprint arXiv:2512.04583},
  year={2025}
}

@article{chen2025high,
  title={High-dimensional tensor discriminant analysis: Low-rank discriminant structure, representation synergy, and theoretical guarantees},
  author={Chen, Elynn and Han, Yuefeng and Li, Jiayu},
  journal={arXiv preprint arXiv:2512.12122},
  year={2025}
}

@article{chen2025modewise,
  title={Modewise additive factor model for matrix time series},
  author={Chen, Elynn and Han, Yuefeng and Li, Jiayu and Xu, Ke},
  journal={arXiv preprint arXiv:2512.25025},
  year={2025}
}

@article{liu2022identification,
  title={Identification and estimation of threshold matrix-variate factor models},
  author={Liu, Xialu and Chen, Elynn Y.},
  journal={Scandinavian Journal of Statistics},
  volume={49},
  number={3},
  pages={1383--1417},
  year={2022},
  publisher={Wiley Online Library}
}

@article{TensorTrain,
  author={Oseledets, I. V.},
  title={Tensor-Train decomposition},
  journal={SIAM Journal on Scientific Computing},
  volume={33},
  number={5},
  pages={2295--2317},
  year={2011},
  doi={10.1137/090752286}
}

@article{zhou2013tensor,
  title={Tensor regression with applications in neuroimaging data analysis},
  author={Zhou, Hua and Li, Lexin and Zhu, Hongtu},
  journal={Journal of the American Statistical Association},
  volume={108},
  number={502},
  pages={540--552},
  year={2013},
  publisher={Taylor \& Francis}
}

@article{li2018tucker,
  title={Tucker tensor regression and neuroimaging analysis},
  author={Li, Xiaoshan and Xu, Da and Zhou, Hua and Li, Lexin},
  journal={Statistics in Biosciences},
  volume={10},
  number={3},
  pages={520--545},
  year={2018},
  publisher={Springer}
}

@article{si2022efficient,
  title={An efficient tensor regression for high-dimensional data},
  author={Si, Yuefeng and Zhang, Yingying and Li, Guodong},
  journal={arXiv preprint arXiv:2205.13734},
  year={2022}
}

@article{li2017parsimonious,
  title={Parsimonious tensor response regression},
  author={Li, Lexin and Zhang, Xin},
  journal={Journal of the American Statistical Association},
  volume={112},
  number={519},
  pages={1131--1146},
  year={2017},
  publisher={Taylor \& Francis}
}

@article{lock2018tensor,
  title={Tensor-on-tensor regression},
  author={Lock, Eric F.},
  journal={Journal of Computational and Graphical Statistics},
  volume={27},
  number={3},
  pages={638--647},
  year={2018},
  publisher={Taylor \& Francis}
}

@article{raskutti2019convex,
  title={Convex regularization for high-dimensional multiresponse tensor regression},
  author={Raskutti, Garvesh and Yuan, Ming and Chen, Han},
  journal={Annals of Statistics},
  volume={47},
  number={3},
  pages={1554--1584},
  year={2019},
  publisher={Institute of Mathematical Statistics}
}

@article{guhaniyogi2017bayesian,
  title={Bayesian tensor regression},
  author={Guhaniyogi, Rajarshi and Qamar, Shaan and Dunson, David B.},
  journal={Journal of Machine Learning Research},
  volume={18},
  pages={1--31},
  year={2017},
  note={Paper No. 79}
}

@inproceedings{he2018boosted,
  title={Boosted sparse and low-rank tensor regression},
  author={He, Lifang and Chen, Kun and Xu, Wanwan and Zhou, Jiayu and Wang, Fei},
  booktitle={Advances in Neural Information Processing Systems 31 (NeurIPS)},
  year={2018}
}

@article{hao2020sparse,
  title={Sparse and low-rank tensor estimation via cubic sketchings},
  author={Hao, Botao and Zhang, Anru R. and Cheng, Guang},
  journal={IEEE Transactions on Information Theory},
  volume={66},
  number={9},
  pages={5927--5964},
  year={2020},
  publisher={IEEE}
}

@article{sun2017provable,
  title={Provable sparse tensor decomposition},
  author={Sun, Will Wei and Lu, Junwei and Liu, Han and Cheng, Guang},
  journal={Journal of the Royal Statistical Society, Series B},
  volume={79},
  number={3},
  pages={899--916},
  year={2017},
  publisher={Wiley}
}

@article{ahmed2020tensor,
  title={Tensor regression using low-rank and sparse {T}ucker decompositions},
  author={Ahmed, Talal and Raja, Haroon and Bajwa, Waheed U.},
  journal={SIAM Journal on Mathematics of Data Science},
  volume={2},
  number={4},
  pages={944--966},
  year={2020},
  publisher={SIAM}
}

@article{cai2023generalized,
  title={Generalized low-rank plus sparse tensor estimation by fast {R}iemannian optimization},
  author={Cai, Jian-Feng and Li, Jingyang and Xia, Dong},
  journal={Journal of the American Statistical Association},
  volume={118},
  number={544},
  pages={2588--2604},
  year={2023},
  publisher={Taylor \& Francis}
}

@article{han2022optimal,
  title={An optimal statistical and computational framework for generalized tensor estimation},
  author={Han, Rungang and Willett, Rebecca and Zhang, Anru R.},
  journal={Annals of Statistics},
  volume={50},
  number={1},
  pages={1--29},
  year={2022},
  publisher={Institute of Mathematical Statistics}
}

@article{zhang2018tensor,
  title={Tensor {SVD}: Statistical and computational limits},
  author={Zhang, Anru and Xia, Dong},
  journal={IEEE Transactions on Information Theory},
  volume={64},
  number={11},
  pages={7311--7338},
  year={2018},
  publisher={IEEE}
}

@article{fan2023factor,
  author={Fan, Jianqing and Gu, Yihong},
  title={Factor augmented sparse throughput deep {ReLU} neural networks for high dimensional regression},
  journal={Journal of the American Statistical Association},
  volume={0},
  number={0},
  pages={1--15},
  year={2023},
  publisher={Taylor \& Francis}
}

@article{schmidthieber2020nonparametric,
  title={Nonparametric regression using deep neural networks with {ReLU} activation function},
  author={Schmidt-Hieber, Johannes},
  journal={Annals of Statistics},
  volume={48},
  number={4},
  pages={1875--1897},
  year={2020},
  publisher={Institute of Mathematical Statistics}
}

@article{bauer2019deep,
  title={On deep learning as a remedy for the curse of dimensionality in nonparametric regression},
  author={Bauer, Benedikt and Kohler, Michael},
  journal={Annals of Statistics},
  volume={47},
  number={4},
  pages={2261--2285},
  year={2019},
  publisher={Institute of Mathematical Statistics}
}

@article{kohler2021rate,
  title={On the rate of convergence of fully connected deep neural network regression estimates},
  author={Kohler, Michael and Langer, Sophie},
  journal={The Annals of Statistics},
  volume={49},
  number={4},
  pages={2231--2249},
  year={2021},
  publisher={Institute of Mathematical Statistics}
}

@inproceedings{novikov2015tensorizing,
  title={Tensorizing neural networks},
  author={Novikov, Alexander and Podoprikhin, Dmitry and Osokin, Anton and Vetrov, Dmitry P.},
  booktitle={Advances in Neural Information Processing Systems 28 (NeurIPS)},
  pages={442--450},
  year={2015}
}

@article{kossaifi2020tensor,
  title={Tensor regression networks},
  author={Kossaifi, Jean and Lipton, Zachary C. and Kolbeinsson, Arinbjorn and Khanna, Aran and Furlanello, Tommaso and Anandkumar, Anima},
  journal={Journal of Machine Learning Research},
  volume={21},
  number={123},
  pages={1--21},
  year={2020}
}

@book{vovk2005algorithmic,
  title={Algorithmic Learning in a Random World},
  author={Vovk, Vladimir and Gammerman, Alex and Shafer, Glenn},
  publisher={Springer},
  year={2005}
}

@article{shafer2008tutorial,
  title={A tutorial on conformal prediction},
  author={Shafer, Glenn and Vovk, Vladimir},
  journal={Journal of Machine Learning Research},
  volume={9},
  number={3},
  year={2008}
}

@article{lei2018distribution,
  title={Distribution-free predictive inference for regression},
  author={Lei, Jing and G'Sell, Max and Rinaldo, Alessandro and Tibshirani, Ryan J. and Wasserman, Larry},
  journal={Journal of the American Statistical Association},
  volume={113},
  number={523},
  pages={1094--1111},
  year={2018},
  publisher={Taylor \& Francis}
}

@inproceedings{romano2019conformalized,
  title={Conformalized quantile regression},
  author={Romano, Yaniv and Patterson, Evan and Cand{\`e}s, Emmanuel J.},
  booktitle={Advances in Neural Information Processing Systems 32 (NeurIPS)},
  year={2019}
}

@article{guan2023localized,
  title={Localized conformal prediction: A generalized inference framework for conformal prediction},
  author={Guan, Leying},
  journal={Biometrika},
  volume={110},
  number={1},
  pages={33--50},
  year={2023},
  publisher={Oxford University Press}
}

@article{zheng2024quantifying,
  title={Quantifying uncertainty in classification performance: {ROC} confidence bands using conformal prediction},
  author={Zheng, Zheshi and Yang, Bo and Song, Peter},
  journal={arXiv preprint arXiv:2405.12953},
  year={2024}
}

@article{zheng2025classification,
  title={Classification uncertainty quantification: A comparison between bootstrap and conformal {ROC} confidence bands},
  author={Zheng, Zheshi and Yang, Bo and Song, Peter},
  journal={Statistica Sinica},
  volume={37},
  number={4},
  year={2025},
  note={In press},
  doi={10.5705/ss.202025.0127}
}

@article{fawcett2006introduction,
  title={An introduction to {ROC} analysis},
  author={Fawcett, Tom},
  journal={Pattern Recognition Letters},
  volume={27},
  number={8},
  pages={861--874},
  year={2006},
  publisher={Elsevier}
}

@book{nakas2023roc,
  title={{ROC} Analysis for Classification and Prediction in Practice},
  author={Nakas, Christos T. and Bantis, Leonidas E. and Gatsonis, Constantine A.},
  year={2023},
  publisher={Chapman and Hall/CRC}
}

@article{majnik2013roc,
  title={{ROC} analysis of classifiers in machine learning: A survey},
  author={Majnik, Matja{\v{z}} and Bosni{\'c}, Zoran},
  journal={Intelligent Data Analysis},
  volume={17},
  number={3},
  pages={531--558},
  year={2013},
  publisher={SAGE Publications}
}

@article{everson2006multi,
  title={Multi-class {ROC} analysis from a multi-objective optimisation perspective},
  author={Everson, Richard M. and Fieldsend, Jonathan E.},
  journal={Pattern Recognition Letters},
  volume={27},
  number={8},
  pages={918--927},
  year={2006},
  publisher={Elsevier}
}

@article{delong1988comparing,
  title={Comparing the areas under two or more correlated receiver operating characteristic curves: A nonparametric approach},
  author={DeLong, Elizabeth R. and DeLong, David M. and Clarke-Pearson, Daniel L.},
  journal={Biometrics},
  volume={44},
  number={3},
  pages={837--845},
  year={1988},
  publisher={JSTOR}
}

@article{venkatraman1996distribution,
  title={A distribution-free procedure for comparing receiver operating characteristic curves from a paired experiment},
  author={Venkatraman, E. S. and Begg, Colin B.},
  journal={Biometrika},
  volume={83},
  number={4},
  pages={835--848},
  year={1996},
  publisher={Oxford University Press}
}

@inproceedings{gyorfi2019nearest,
  title={Nearest neighbor based conformal prediction},
  author={Gy{\"o}rfi, Laszlo and Walk, Harro},
  booktitle={Annales de l'ISUP},
  volume={63},
  number={2-3},
  pages={173--190},
  year={2019}
}

@misc{anandkumar2015,
  title={Guaranteed non-orthogonal tensor decomposition via alternating rank-1 updates},
  author={Anandkumar, Animashree and Ge, Rong and Janzamin, Majid},
  year={2015},
  eprint={1402.5180},
  archivePrefix={arXiv},
  primaryClass={cs.LG},
  url={https://arxiv.org/abs/1402.5180}
}

%
%
%

\end{document}